\documentclass[preprint,12pt]{elsarticle}

\usepackage{amssymb}
\usepackage{amsmath}
\usepackage{svg}
\usepackage{colortbl}
\usepackage[margin=1in]{geometry}
\usepackage[frozencache=true,cachedir=_minted-main]{minted} 
\usepackage{subcaption}
\usepackage{bm}
\usepackage{multirow}
\usepackage{xurl}
\usepackage{url}

\journal{Knowledge-Based Systems}

\begin{document}

\begin{frontmatter}



\title{Application of machine learning to monster level prediction in tabletop RPG game design}

\author[agh]{Jolanta Śliwa\fnref{eq}}
\ead{jolantasliwa@agh.edu.pl}

\author[agh]{Jakub Adamczyk\corref{cor}\fnref{eq}}
\ead{jadamczy@agh.edu.pl}

\affiliation[agh]{organization={Faculty of Computer Science, AGH University of Krakow},
            city={Cracow},
            country={Poland}
}

\cortext[cor]{Corresponding author}

\fntext[eq]{Both authors contributed equally to this work.}
\fntext[label1]{ORCID 0009-0004-1889-5655}
\fntext[label2]{ORCID 0000-0003-4336-4288}

\begin{abstract}
Designing balanced adversaries is a central but labor-intensive task in tabletop role-playing game (TTRPG) development. In systems such as Pathfinder, each monster is described by many numerical attributes that jointly determine its power, summarized as an ordinal level. We investigate whether machine learning can support designers by predicting this level from a monster's attributes, framing the task as tabular ordinal regression. We introduce what is, to our knowledge, the first dataset built specifically for TTRPG monster-level prediction, derived from publicly available Pathfinder Second Edition data. Using it, we compare classical regression models with rounding schemes, dedicated tabular ordinal regression algorithms, and neural networks with ordinal-aware losses. To mirror real design workflows, we evaluate all models under chronological and expanding-window protocols with several complementary metrics. Results show that tree-based ensembles outperform linear models and neural approaches, achieving near-perfect ordinal ranking and high predictive accuracy. Explainable AI analyses, such as feature importance and error distributions, show that the model is aligned with human intuition and follows patterns grounded in game rules. Together, these results show that machine learning can reliably approximate designer judgments and serve as an effective computer-aided tool for monster balancing and broader TTRPG system design.
\end{abstract}


\begin{keyword}
machine learning \sep ordinal regression \sep computer-aided design \sep game design

\MSC 62-04 \sep 62J02 \sep 62P25 \sep 68-04 \sep 68T05 \sep 68T35 \sep 68U07
\end{keyword}

\end{frontmatter}



\section{Introduction}

Artificial intelligence (AI) and machine learning (ML) methods have been used to play games since the inception of these fields, with breakthroughs such as alpha-beta pruning \cite{alpha_beta_pruning}, Monte Carlo Tree Search (MCTS) \cite{MCTS}, AlphaGo \cite{AlphaGo}, and AlphaZero \cite{AlphaZero}. These applications are player-centered and assume an existing game environment together with its rules. In other words, they presuppose that the game itself has already been created and designed, which allows an abstract mathematical representation of its rules and states. In this work, we instead focus on using machine learning for \textit{game design}, an area that has attracted comparatively less interest yet is of high importance to the industry, as many steps of game creation are slow, labor-intensive, and expensive.

Specifically, we study role-playing games (RPGs), and in particular pen \& paper RPGs, also known as \textit{tabletop RPGs (TTRPGs)}. These are distinct from, and predate, computer games and computer RPGs (cRPGs). In this type of game, participants take on the roles of fictional characters within a story narrated by a Game Master (GM). Players gather around a table, physically or virtually, to create and develop characters within an evolving story. Gameplay is driven by narration and by the resolution of conflicts through a chosen \textit{rule system} (or simply ``system''), typically using multi-sided dice rolls to introduce randomness. The game design task considered here centers purely on the rules of the game itself, rather than on generating story, levels, or music, which are left to the GM and human creativity. TTRPGs are also frequently adopted as rulesets for computer games, with famous examples such as the Baldur's Gate series using Dungeons \& Dragons (D\&D) rules, and Pathfinder: Wrath of the Righteous using the Pathfinder First Edition system.

The RPG market has been growing steadily for a long time. In 2023, it was valued at approximately $1.92$ billion USD, with projections estimating that it will reach $5.27$ billion USD by 2033 \cite{ttrpg-growth}. This rapid expansion highlights the increasing popularity of TTRPGs and the evolving demands of players. TTRPGs offer advantages over computer RPGs, which have a fixed narrative and leave players limited freedom, compared with the virtually infinite possibilities of TTRPGs. However, to remain competitive, publishers, miniature manufacturers, and other industry players must continuously innovate. Beyond creative marketing strategies and engaging gameplay demonstrations, companies place a strong emphasis on R\&D and technological advancement to improve the gaming experience. There is growing interest in applying AI and ML methods in this area, and some aspects are particularly well suited for them.

One of the most important components of adventures is combat with adversaries, commonly known as enemies or \textit{monsters}, introduced by the GM, such as goblins or dragons. Like player characters, monsters are represented in the rule system by numerous numerical \textit{attributes} that define their strength, abilities, and difficulty to defeat. Examples include strength, the amount of damage dealt by each attack, and armor. In many systems, the overall power of a monster is indicated by a value called its \textit{level}, typically a non-negative integer, with higher levels indicating more powerful monsters. The level is directly and causally derived from the attribute values, as monsters with high values have higher levels and are designed to be encountered by more powerful players. Selecting monsters of an appropriate level is essential for preparing encounters that are challenging yet winnable for players.

An important task for an RPG publisher is creating whole books of example monsters that a GM can use in their own narrative. This requires properly \textit{estimating the level of each monster from its statistics}. However, in many systems, such as D\&D or Pathfinder, monsters are described by dozens of attributes, which makes this process hard for human designers. In practice, such estimation is done either through expert human judgment, which is error-prone, or through many hours of gameplay, which is labor-intensive and typically requires 4-5 people (a GM and players) at a time.

Selling books of monsters constitutes an important source of income for these companies, and a single compendium can contain over 100 monsters. This creates a business need to quickly and automatically estimate a monster's level from its numerical statistics, obtaining early validation of potential designs. ML-based solutions lend themselves naturally to this problem, as they can readily incorporate high-dimensional numerical inputs and non-linear relationships between attributes and level. Such tools can help game designers accelerate the process by providing feedback on monster power during creation. Furthermore, GMs can use these algorithms to design unique monsters, enhancing their own gameplay experience.

\textbf{Key contributions} of this work include:
\begin{itemize}
    \item Creation of what is, to our knowledge, the first dataset for TTRPG game design, specifically for monster level estimation in the Pathfinder Second Edition game system. It is publicly available under a permissive open source license, to foster further research in this area.
    \item Formalization of monster level prediction as a tabular ordinal regression problem. We provide feature engineering examples and perform a comprehensive benchmark of 16 models for this task, including regression with rounding, dedicated tabular ordinal regression algorithms, and neural networks with ordinal loss functions.
    \item Domain-specific evaluation schemes for a realistic quantification of model generalization, including a time-based expanding-window train-test split and metrics such as macro-averaged MAE for imbalanced regression.
    \item Application of data analysis and explainable AI (XAI), demonstrating notable robustness of the models, alignment with human intuition, and real-world applicability.
\end{itemize}

The created dataset and all code needed to reproduce the experiments and results are publicly available on GitHub, under permissive open source licenses: \url{https://github.com/tunczyk101/Monster-level-prediction-in-TTRPG}.

\section{Literature review}

AI and ML have been widely applied in games, but primarily for playing them, particularly through search techniques and reinforcement learning. The area addressed in this work, game design, has attracted much less research interest.

Much of the existing literature is concerned more with humanities aspects than with computer science, for example storytelling \cite{drachen2009towards} or the integration of technology to improve traditional gameplay experiences \cite{tychsen2006making}. Applications of generative ML have been explored in procedural world design and creative content generation, such as map generation \cite{map_generation,map_generation_2,map_generation_3,map_generation_4,map_generation_5}, music generation \cite{music_generation,music_generation_2}, and non-player character (NPC) text and conversation generation \cite{text_generation,text_generation_2,text_generation_3}.

Predictive AI and ML methods, in particular for gameplay balance, have been much less explored. The closest work to ours is adaptive gameplay balancing \cite{adaptive_game_design}, which dynamically adjusts the game world, such as enemy statistics, based on player behavior. However, these techniques adaptively control player adversaries, which means that they apply only to computer games, operate on an already existing game system rather than being part of game design, and require deep integration into the software. As such, they fall outside the scope of this work, which is concerned with predicting monster level as part of game design, before gameplay begins.

Several reviews of AI and ML applications in games have been published in recent years \cite{ai_in_games_review,ai_in_games_review_2,ai_in_games_review_3,ai_in_games_review_4,ai_in_games_review_5,ai_in_games_review_6}. However, they focus heavily on the AI-based content generation and ML-based interactive gameplay described above, and none even mention the possibility of using predictive ML for enemy power prediction or statistic balancing. To our knowledge, this work is therefore the first to propose such an application.

The monster level in RPG games is typically an integer starting from 0 or 1. Consequently, when predictive ML methods are applied, the dependent variable is both discrete and ordered, which constitutes an \textit{ordinal regression} problem, also known as ordinal classification \cite{winship1984regression, burkner2019ordinal, ord_reg_review}. Various algorithms have been proposed for this problem, which arises in many ML applications, for example recommender systems (user ratings such as 1-5 stars) \cite{ord_reg_collaborative_filtering}, age estimation (predicting the age of a subject in an image) \cite{niu2016ordinal}, and strength of opinion in the social sciences and psychology (a value on a Likert scale) \cite{burkner2019ordinal}. We could not find any prior application of ordinal regression to games or game design; this too constitutes a novel contribution of our work.

One group of methods is based on standard regression, with the results rounded as a post-processing step to ensure discrete outputs \cite{kramer2001prediction, shin2022moving}. Various rounding schemes can be used in place of classical mathematical rounding to improve results. However, this approach does not incorporate the information of discrete classes into the learning process. Alternatively, many classical tabular ML models have been adapted for ordinal regression, including ordinal logistic regression \cite{ord}, Ordered Random Forest (ORF) \cite{orf}, and Gaussian Process Ordinal Regression (GPOR) \cite{gpor}. For neural networks, dedicated layers and loss functions for ordinal regression have been proposed, such as CORN \cite{corn}, CORAL \cite{coral2020}, and CONDOR \cite{condor}. Detailed descriptions of these algorithms are presented in the Methods section.

\section{Dataset and feature engineering}
\label{sec:dataset}

In this section, we describe the dataset construction and its basic characteristics. We also explain the approach taken to feature engineering, in order to obtain the tabular dataset used for downstream model training.

\subsection{Dataset curation}

In this work, we focus on the Pathfinder Roleplaying Second Edition game system (referred to as \textit{Pathfinder 2e} for short) for three reasons. First, it is highly popular, consistently ranking among the most popular TTRPGs, which makes ML solutions of high interest to many potential users. Second, it is highly mathematical: monsters are described by roughly a dozen basic attributes and many additional ones, which makes estimating their level challenging for humans yet provides ample opportunity for feature engineering. Third, it is published under permissive open source licenses (older works under the Open Game License 1.0A \cite{paizo-pathfinder-license}, newer ones under the Open RPG Creative (ORC) License \cite{pathfinder-orc}), which enables open science principles and allows us to share the data for reproducibility and further research.


We created a dataset of monsters based on a publicly available database of Pathfinder 2e books, amounting to 6007 monster entries in total from 302 sourcebooks, adventures, and scenarios, the newest of which was published on 26.04.2026. The full list of source books used is available in the GitHub repository. Creatures come from books published by Paizo; each book provides either many monsters, in the case of monster compendia, or only a few unique ones, in the case of single adventure scenarios. We note the strong temporal characteristics of monster design: newer monsters are designed based on prior ones. For example, a novel yet similar variant of an existing monster should have a level comparable to that of the existing one.

\begin{figure}
    \centering
    \includegraphics[width=0.75\linewidth]{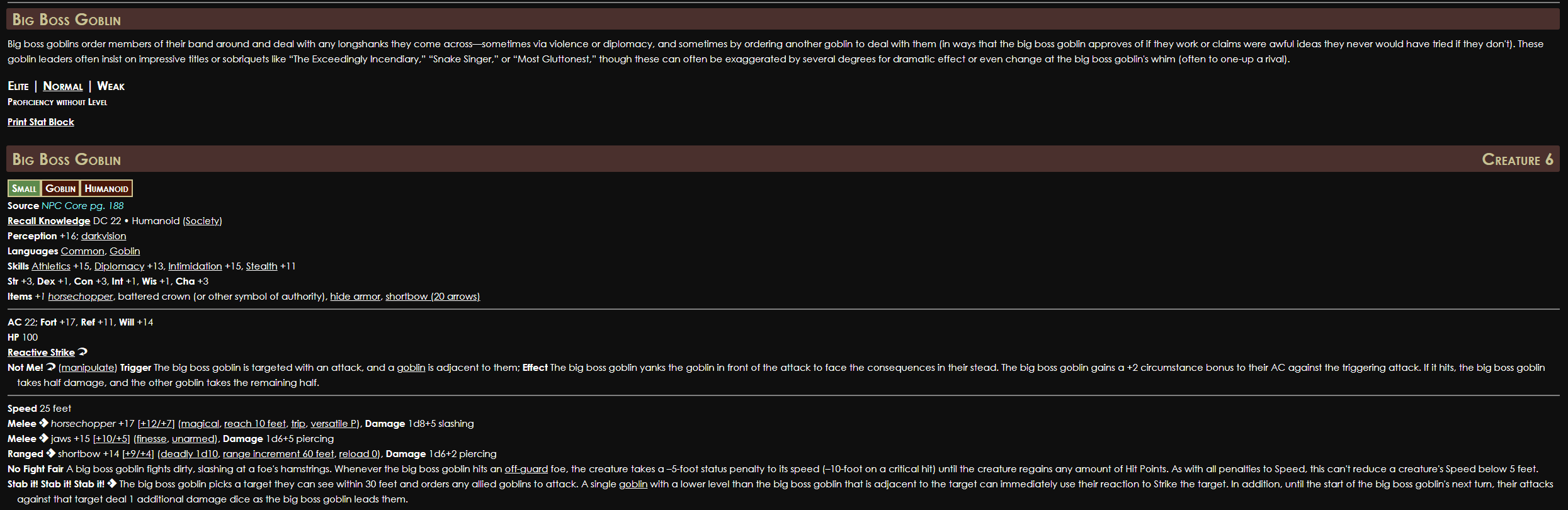}
    \caption{A monster example from Pathfinder 2e \cite{nethys-monster}.}
    \label{fig:monster}
\end{figure}

A monster entry provides descriptions and characteristics as a highly complex JSON object. A rendered display of a monster statistics block, from the Archives of Nethys web page (officially supported by the publisher Paizo) \cite{archives_of_nethys}, is provided in Figure \ref{fig:monster}.

\begin{figure}[htp]
    \centering
    \includegraphics[width=0.75\textwidth]{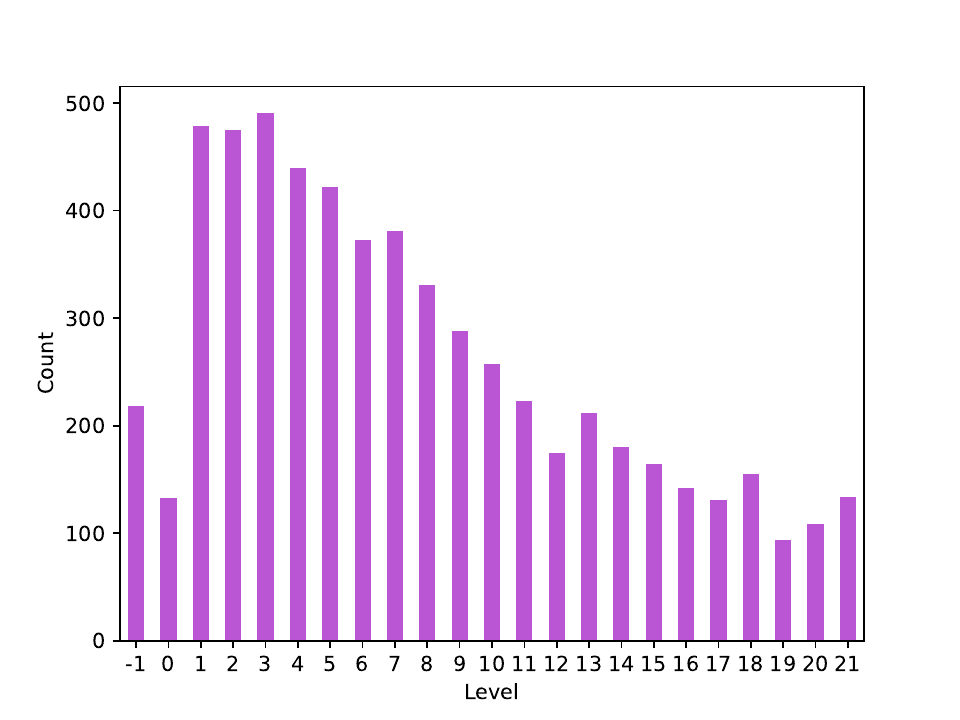}
    \caption{Distribution of monsters' levels.}
    \label{fig:levels}
\end{figure}

In particular, each entry contains the monster level, which we treat as the ground truth. This level was obtained through expert estimation and extensive practical gameplay. In the Pathfinder 2e system, the level is an integer ranging from -1 to 25. Because very few monsters exceed level 20, with only 89 in total across levels 21-25, we aggregate them into a single category denoting ``level over 20''; for convenience, we mark it simply as 21. The distribution of this dependent variable is shown in Figure \ref{fig:levels}. This is an \textit{imbalanced regression} problem, with comparatively few monsters having very low levels (-1 or 0) and very high levels (19-21).

Level is an ordinal variable, being both ordered and discrete, but it does not lie on a ratio scale. For example, a level 4 monster is not necessarily twice as powerful as a level 2 monster, as higher-level monsters often possess unique abilities or powerful spells. The problem therefore requires learning this ordered relation, which is more sophisticated than merely regressing on integers.

We also include an important metadata variable: the source book in which each monster was originally published. By checking the publication date of each book, we can incorporate chronological information about the order in which monsters were designed. As we show in the Experiments and results section, this is crucial for a realistic evaluation of ML models.

\subsection{Feature engineering}

The raw input data is complex and contains a significant amount of information that is not directly relevant to predicting the monster level, and in particular information that would not be expected to be available from users at prediction time. We therefore focus our dataset curation and feature engineering on attributes known to domain experts to be directly and causally related to the level. For example, a monster with very high armor must have a high level, since otherwise it would be impossible for low-level players to defeat. This yields a tabular ML problem in which each monster is described by a vector of features. We provide three feature sets extracted from the raw statistics: basic, expanded, and full.

The \textit{basic} set consists of 8 fundamental statistics shared by all monsters, as well as by players, which influence almost every aspect of the Pathfinder system. The first 6 are the basic attributes, from which many other characteristics are derived: Strength (Str), Dexterity (Dex), Constitution (Con), Intelligence (Int), Wisdom (Wis), and Charisma (Cha). We also include Hit Points (HP), which measure the health or durability of the monster, and Armor Class (AC), which indicates how hard the monster is to hit with an attack. Both directly influence the survivability of the adversary and are therefore closely related to the level.

The \textit{expanded} set adds 12 further characteristics: 4 related to defensive abilities, 4 to offensive, and 4 determining its magical power.

Perception measures how easily a monster detects players, and also captures some special abilities. We further include three so-called saving throws, which determine resistance to spells, special abilities, and attacks: Fortitude, Reflex, and Will. This set also includes attack features, as nearly every monster possesses at least one melee and one ranged attack, since their main role is to fight players. In the Pathfinder 2e system, an attack involves rolling a 20-sided die (d20) and modifying the result with a so-called attack bonus. If the value is at least equal to the target's Armor Class (AC), the attack succeeds and damage is calculated. Damage uses one or more multi-sided dice plus a damage modifier. Monsters may possess several attacks of each type, but most often use the one with the highest attack bonus. For each monster, we therefore extracted the highest available attack bonus for both melee and ranged attacks, and computed the expected damage of the corresponding attack, which resulted in 4 features.

For example, for an attack described as \textit{``2d8 + 1d4 + 3''}, the expected damage is the sum of the expected values of each die roll plus the modifier. Here \textit{d8} denotes an 8-sided die, which forms a discrete uniform distribution over the values $[1, 2, \dots, 8]$, i.e., $\mathcal{U}(1,8)$. With a slight abuse of expected-value notation, the expected damage is:

$$
\mathbb{E}[2d8 + 1d4 + 3] = 2 \cdot \mathbb{E}[\mathcal{U}(1,8)] + \mathbb{E}[\mathcal{U}(1,4)] + 3 = 14.5
$$

Some monsters are mages, able to cast spells according to their power. The Pathfinder system divides spells into 9 spell levels based on their power. We include the highest available spell level as a feature, because characters (including players) gain access to higher-level spells as they advance, so there is a strong correlation between available spell level and monster level. Furthermore, mages of the same spell level can differ in power, which is captured by two characteristics: the spell attack modifier, used to hit targets with spells, and the spell Difficulty Class (DC), which determines how challenging it is for a target to resist a spell cast by the monster. We therefore include both of these features as well. There is also a type of spells that require special points, called Focus. Even classes that are not typical spellcasters can use them, and because of that, Focus was the last feature included as part of this set.

The \textit{full} set adds 13 more complex characteristics, in particular several derived from the basic features through complex, non-linear formulas. Many of these features are optional; for example, most monsters cannot fly and therefore have a fly speed of zero. We include features related to movement and to immunities against damage and magic spells.

First, we add 3 speed features: land, fly, and swim speed. These also provide implicit information about the availability of powerful movement types; for example, $fly > 0$ means that the monster can fly.
 
Next, we add the total number of a monster's immunities and resistances to certain damage types, such as a fire dragon being immune to fire. Because immunity types are numerous (over a dozen) and very sparse (most monsters have none, or at most one or two), we include the total number of immunities as a single feature. Initial experiments showed that this approach yields performance similar to or better than using separate features.
 
In addition to the spell-related features above, we include the number of spells available at each level as 9 separate features. This is a more detailed information than just the highest spell level. Introducing both improved the results in initial experiments.

In principle, models could learn the maximum spell level implicitly from the available spells at each level. However, most likely because of the limited amount of data, explicitly including the maximum spell level alongside the 9 per-level spell-count features further improved results in initial experiments.

We summarize the three feature sets in Table \ref{tab:features}. Gains from additional features, together with feature importance analyses, are detailed in the Experiments and results section.

\begin{table}[]
\centering
\caption{Summary of the three feature sets.}
\resizebox{\textwidth}{!}{%
\begin{tabular}{|c|c|c|}
\hline
\textbf{Feature set} & \textbf{\# of features} & \textbf{Description} \\ \hline
Basic & 8 & Attributes (6), HP, AC \\ \hline
Expanded & 20 & Perception, saving throws (3), attacks (4), spell attacks and focus (4) \\ \hline
Full & 33 & Movement (3), number of immunities, spells per level (9) \\ \hline
\end{tabular}
\label{tab:features}
}
\end{table}

\section{Methods}

We frame the problem of predicting the monster level as tabular ordinal regression. In ordinal regression, also known as ordinal classification, the dependent variable is discrete and ordered, with target values (classes) typically represented as integers. However, the intervals between categories are not necessarily equal. Moreover, evaluation must take the error magnitude into account, since predicting class 2 instead of class 1 is less problematic than predicting class 3. These characteristics make ordinal regression similar to both regression and classification, yet distinct in its own right.

The simplest approach is to treat the problem as standard regression and round the predictions as a post-processing step. This assumes a ratio scale and introduces rounding error, but it is simple and allows the use of arbitrary regression models. Alternatively, many ML models have been modified to incorporate inductive biases specific to ordinal regression. An important such bias is a cumulative-probability (rank-consistency) guarantee: the model should predict monotonically non-increasing scores for higher classes, i.e., a level 4 monster is at least as likely to be ``at least level $k$'' as a level 3 monster. Similar ideas can also be used to construct loss functions for neural networks.

\subsection{Notation}

To keep the presentation uniform, we use the following notation throughout this section. The training set is $\{(\bm{x}_i, y_i)\}_{i=1}^{N}$ of $N$ monsters, where $\bm{x}_i \in \mathbb{R}^d$ is the feature vector of the $i$-th monster and $y_i \in \{1, \dots, K\}$ is its ordinal level, one of $K$ ordered classes. The index $i$ always denotes a \emph{sample}, and the index $k$ always denotes a \emph{class} or, equivalently, one of the $K-1$ ordinal \emph{thresholds} that separate consecutive classes.

Many methods reduce ordinal regression to $K-1$ binary subproblems, in which the $k$-th subproblem asks whether $y_i > k$; we write $f_k(\bm{x})$ for the (probabilistic) output of the $k$-th binary sub-model and $f(\bm{x})$ for a single scalar score. Thresholds (also called cutpoints) are written $\theta_1 < \theta_2 < \dots < \theta_{K-1}$, with the convention $\theta_0 = -\infty$ and $\theta_K = +\infty$.

We denote by $\bm{w}$ a vector of linear weights, by $\sigma(\cdot)$ the logistic sigmoid, and by $\Phi(\cdot)$ the standard normal cumulative distribution function (the probit function). Estimated quantities carry a hat, for example $\hat{y}_i$ for the predicted level and $\hat{P}(y_i = k)$ for a predicted class probability. Neural network parameters are collectively denoted $\Theta$.

\subsection{Regression models with rounding}

Tabular regression models, both linear and non-linear, can be applied directly to monster level prediction. These include linear regression with ridge (L2) regularization (Ridge), support vector machines (SVM) \cite{SVM}, Random Forest \cite{random_forest}, and boosting methods such as LightGBM \cite{LightGBM}.
 
An approach based on k-nearest neighbors (kNN) is implicitly described in the Pathfinder 2e rulebook, which mentions estimating a monster's level by comparison with the most similar monsters in official books. kNN is therefore an important baseline that reflects human-level performance.
 
After applying such a method, the continuous output is transformed into a discrete ordinal label by \textit{rounding}. Many options exist here, from mathematical rounding (a threshold of 0.5 between levels) to sophisticated stacking using an additional classifier model \cite{regression-to-build-classifiers}. In this work, we use mathematical rounding, which maps continuous predictions up or down based on a fixed threshold of 0.5. It introduces no additional parameters, albeit it may give suboptimal results for imbalanced regression. For example, if the model consistently predicts values that are too small, it could be beneficial to use a lower threshold, such as rounding up at a threshold of 0.3. We leave the exploration of this idea for future work.

\subsection{Tabular ordinal regression models}

We selected several dedicated tabular ordinal regression models for evaluation. In particular, we focused on modified variants of the regression models above, for example Ordered Random Forest \cite{orf}, to compare with Random Forest regression with rounding. This lets us fairly evaluate the gains from these problem-specific modifications. Selection also relied on the availability of easy-to-use, open source implementations in Python.

\subsubsection{Simple Approach to Ordinal Classification (SAOC)}

Frank and Hall \cite{frank2001simple} proposed a simple, universal way to adapt an arbitrary binary classifier to ordinal regression, which we call the Simple Approach to Ordinal Classification (SAOC). The original problem with $K$ ordered classes is decomposed into $K-1$ binary classification problems, indexed by $k \in \{1, \dots, K-1\}$, and one binary classifier is trained for each. The $k$-th classifier estimates the probability that the level exceeds $k$, i.e., $f_k(\bm{x}) \approx P(y > k \mid \bm{x})$. For the $k$-th problem, the label of sample $i$ is transformed according to Equation \ref{equation:saoc_transform}.

\begin{equation}
y_i^{(k)} =
\begin{cases}
    1 & \text{if } y_i > k \\
    0 & \text{if } y_i \leq k \\
\end{cases}
\label{equation:saoc_transform}
\end{equation}

In particular, for the first classifier ($k = 1$), every sample whose true level is 1 is relabeled 0, and every sample of a higher level is relabeled 1. For the last classifier ($k = K-1$), only samples of the highest level $K$ are relabeled 1, and all others 0. Because of this construction, no classifier with index $K$ is needed.

This yields $K-1$ datasets with the same features but different transformed labels, on each of which we train a standard binary classifier. To make a prediction for a new sample $\bm{x}$, we first obtain $f_k(\bm{x}) = \hat{P}(y > k \mid \bm{x})$ from each classifier and then recombine them into class probabilities following Equation \ref{equation:saoc_prediction}.

\begin{equation}
\hat{P}(y = k \mid \bm{x}) =
\begin{cases}
    1 - f_1(\bm{x}) & \text{if } k = 1 \\
    f_{k-1}(\bm{x}) \, \bigl(1 - f_k(\bm{x})\bigr) & \text{if } 1 < k < K \\
    f_{K-1}(\bm{x}) & \text{if } k = K \\
\end{cases}
\label{equation:saoc_prediction}
\end{equation}

We then predict the class with the highest probability. Because the binary classifiers are trained independently, they need not be \textit{consistent}: a model may assign $f_k(\bm{x}) > f_{k-1}(\bm{x})$, i.e., a higher probability to $y > k$ than to $y > k-1$. This should not happen for a well-trained model with enough data, but it is not formally enforced.

In our experiments, SAOC uses a Random Forest as the underlying binary classifier. It gave stable results in initial experiments, is fast to train and parallelizable (which is highly relevant when training $K-1 = 21$ underlying models), and has low sensitivity to hyperparameter values \cite{random_forest_tunability}.

\subsubsection{Ordered Logistic Regression (ORD)}

Ordered Logistic Regression (ORD) \cite{ord} adapts logistic regression to ordinal labels. It remains a linear model, but replaces the standard odds with cumulative odds, modeling the probability that the level is at most $k$. This has the considerable advantage of guaranteeing consistent predictions by incorporating the ordering into the learning process. Writing $p_k = P(y = k)$, the cumulative probability of class $k$ or below is given by Equation \ref{equation:ord_proba}, the corresponding odds by Equation \ref{equation:ord_odds}, and the logit by Equation \ref{equation:ord_logit}.

\begin{equation}
    P(y \leq k) = p_1 + \dots + p_k
\label{equation:ord_proba}
\end{equation}
 
\begin{equation}
    \mathrm{odds}(y \leq k) = \frac{P(y \leq k)}{P(y > k)} = \frac{p_1 + \dots + p_k}{p_{k + 1} + \dots + p_K}
\label{equation:ord_odds}
\end{equation}
 
\begin{equation}
    \mathrm{logit}(y \leq k) = \ln \frac{P(y \leq k)}{1 - P(y \leq k)}, \quad k = 1, \dots, K - 1
\label{equation:ord_logit}
\end{equation}

This model is based on the proportional-odds assumption, which states that the effect of the predictors is constant across all thresholds. In other words, if a particular feature increases the probability of belonging to a lower class, it does so consistently across all boundaries between classes. The weight vector $\bm{w}$ is shared across thresholds, while only the intercepts (which play the role of the cutpoints $\theta_k$) vary. Under this assumption, the cumulative logit for threshold $k$ is:

\begin{equation}
\mathrm{logit}(y \leq k) = \theta_k - \bm{w}^\top \bm{x}, \quad k = 1, \dots, K-1
\label{equation:ord_cumulative_logit}
\end{equation}

where $\bm{x}$ is the feature vector, $\bm{w}$ the shared regression coefficients, and $\theta_k$ the intercept (cutpoint) of the $k$-th cumulative probability. The coefficients $\bm{w}$ and the cutpoints $\theta_1, \dots, \theta_{K-1}$ are estimated jointly by maximum likelihood.

\subsubsection{Threshold-based linear models}

Threshold-based models \cite{rennie2005loss} were proposed to generalize the loss functions of linear classifiers to multiple ordered categories. The framework generalizes the single-threshold setting of binary classification by introducing $K-1$ thresholds $\theta_1 < \theta_2 < \dots < \theta_{K-1}$ for $K$ classes, together with $\theta_0 = -\infty$ and $\theta_K = +\infty$. These thresholds partition the real line into $K$ segments, and a scalar model output $f(\bm{x})$ that falls in $(\theta_{k-1}, \theta_k)$ assigns the instance to class $k$.

Two loss variants are used: Immediate-Threshold (IT) and All-Threshold (AT). Both generalize an arbitrary binary loss $\ell$, and we use the hinge loss for its robustness to outliers. It can be shown \cite{or_consistency} that IT is a surrogate for the 0-1 loss, while AT is a surrogate for the absolute loss.

The Immediate-Threshold (IT) variant penalizes only the two thresholds that bound the correct class. For a sample $(\bm{x}_i, y_i)$ with $y_i = k$, the pair $(\theta_{k-1}, \theta_k)$ delimits the range in which the output is considered correct, and predictions outside it are penalized, as shown in Equation \ref{equation:it_loss}.

\begin{equation}
    L_{\mathrm{IT}}\bigl(f(\bm{x}_i), y_i\bigr) = \ell\bigl(f(\bm{x}_i) - \theta_{y_i - 1}\bigr) + \ell\bigl(\theta_{y_i} - f(\bm{x}_i)\bigr)
\label{equation:it_loss}
\end{equation}

The All-Threshold (AT) variant sums penalties over \emph{all} thresholds, as shown in Equation \ref{equation:at_loss}. This not only encourages correct classification but also more heavily penalizes predictions far from the correct label. The penalty grows linearly with the distance between the predicted and the correct class.
 
\begin{equation}
    L_{\mathrm{AT}}\bigl(f(\bm{x}_i), y_i\bigr) = \sum_{k=1}^{K-1}
    \begin{cases}
        \ell\bigl(\theta_k - f(\bm{x}_i)\bigr) & \text{if } k \geq y_i \\
        \ell\bigl(f(\bm{x}_i) - \theta_k\bigr) & \text{if } k < y_i
    \end{cases}
\label{equation:at_loss}
\end{equation}

In our initial experiments, the AT variant performed consistently slightly worse than the IT variant. Thus, we report results only for the latter.

\subsubsection{Ordered Random Forest}

The Ordered Random Forest (ORF) \cite{orf,orf2} has been proposed as an adaptation of the Random Forest to ordinal regression. It estimates the conditional class probabilities $P(y_i = k \mid \bm{x}_i)$ from cumulative, nested binary indicators $\mathbf{1}\{y_i \leq k\}$ for $k \in \{1, \dots, K-1\}$. This decomposes the problem into $K-1$ binary classification problems, whose forests produce cumulative estimates (Equation \ref{equation:orf_cum_prob}).

\begin{equation}
    \hat{F}_k(\bm{x}_i) = \hat{P}(y_i \leq k \mid \bm{x}_i)
\label{equation:orf_cum_prob}
\end{equation}

The class probabilities are then obtained as successive differences of the cumulative estimates (Equation \ref{equation:orf_probas})

\begin{equation}
    \hat{P}(y_i = k) =
\begin{cases}
     \hat{F}_1(\bm{x}_i) & \text{if } k = 1 \\
     \hat{F}_k(\bm{x}_i) - \hat{F}_{k-1}(\bm{x}_i) & \text{if } 1 < k < K \\
     1 - \hat{F}_{K-1}(\bm{x}_i) & \text{if } k = K
\end{cases}
\label{equation:orf_probas}
\end{equation}

Because the binary forests are trained independently, a difference can occasionally be negative for imbalanced data with sparsely populated classes. Therefore, we clip each estimate at zero, $\hat{P}(y_i = k) \leftarrow \max\bigl(0, \hat{P}(y_i = k)\bigr)$, and renormalize so that the class probabilities sum to one (Equation \ref{equation:orf_normalization}). In practice this correction is rarely needed for well-performing models.

\begin{equation}
    \hat{P}(y_i = k) = \frac{\hat{P}(y_i = k)}{\sum_{k'=1}^{K} \hat{P}(y_i = k')}, \quad 1 \leq k \leq K
\label{equation:orf_normalization}
\end{equation}

\subsubsection{Gaussian Processes for Ordinal Regression}

Gaussian Processes for Ordinal Regression (GPOR) \cite{gpor} have been proposed to adapt the Gaussian processes (GPs) for ordinal regression tasks. They provide a flexible, nonparametric, kernel-based probabilistic model with built-in uncertainty estimates, which can be an additional benefit for game designers when balancing monsters. In this method, the latent function $f$ is modeled by a Gaussian process, and the likelihood is defined by a threshold model that generalizes the probit function to ordinal targets, interpreting each ordinal label as an interval of the latent space delimited by the thresholds.

Given a latent value $f = f(\bm{x})$, thresholds $\theta_0 < \theta_1 < \dots < \theta_K$ (with $\theta_0 = -\infty$, $\theta_K = +\infty$), and a scale (noise) parameter $s$, the probability of class $k$ is expressed through the standard normal CDF $\Phi(\cdot)$ (Equation \ref{equation:gpor_function}).

\begin{equation}
P(y = k \mid f) = \Phi\!\left(\frac{\theta_k - f}{s}\right) - \Phi\!\left(\frac{\theta_{k-1} - f}{s}\right), \quad k = 1, \dots, K
\label{equation:gpor_function}
\end{equation}

The thresholds $\theta_k$ and the scale $s$ are learned by maximizing the posterior given the observed data $D$ (Equation \ref{equation:gpor_posterior}), where $P(D \mid \theta, s)$ is the data likelihood. A prior may be specified from domain knowledge, chosen to be uninformative, or omitted. The last option reduces the objective to log marginal likelihood maximization (MLE) rather than maximum a posteriori (MAP) estimation.

\begin{equation}
    P(\theta, s \mid D) \propto P(D \mid \theta, s)\, P(\theta)
\label{equation:gpor_posterior}
\end{equation}

This model combines the flexibility of Gaussian processes with an explicit ordinal likelihood. A known limitation is its poor scalability in the number of training samples, on the order of $O(N^3)$. For Pathfinder this is not particularly problematic, as the number of monsters is relatively small. We use the RBF (squared exponential) kernel and set no prior, maximizing the pure log marginal likelihood, since this combination is highly flexible and data-driven. GPflow \cite{gpflow} is used for the implementation, with a Variational Gaussian Process (VGP) as we have a non-Gaussian likelihood, and parameters are optimized using L-BFGS-B.

\subsection{Neural networks for ordinal regression}

The models above introduce modifications that assume a particular functional form of the underlying model. Adapting neural networks to ordinal regression requires more general, differentiable modifications, many of which have been proposed in the literature. Typically, ordinal regression is decomposed into a series of binary classification tasks, with a problem-specific inductive bias introduced through the loss function and/or the final layers of a multilayer perceptron (MLP) classifier.

Because our data is tabular, we focus exclusively on the MLP model. We keep a similar number of layers and parameters across architectures. Our base network has 2 hidden layers (128 and 64 neurons, respectively) with ReLU activation, followed by a model-dependent output layer whose number of outputs, activation, and specific modifications depend on the architecture, as described below. These choices were made based on initial experiments, which showed no significant gains from larger or more heavily regularized models. In what follows, $\Theta$ denotes the network parameters.

\subsubsection{Cumulative link model}

The cumulative link model (CLM) \cite{clm,or_consistency} takes a probabilistic view of the problem: instead of scoring each class directly, it models the \emph{cumulative} class probabilities through a link function. It belongs to the same family of threshold-based surrogates, like IT and AT losses described above, but replaces the margin loss with a negative log-likelihood. In our case the scalar score $f(\bm{x})$ is produced by an MLP with parameters $\Theta$.

Given the score $f(\bm{x})$, ordered cutpoints (thresholds) $\theta_1 \leq \theta_2 \leq \dots \leq \theta_{K-1}$, and a link function $\sigma$, the CLM sets the cumulative probability of class $k$ or below to:

\begin{equation}
\sigma\bigl(\theta_k - f(\bm{x})\bigr) = \hat{P}(y \leq k \mid \bm{x}), \qquad k = 1, \dots, K-1
\label{equation:clm_cumulative}
\end{equation}

Using the logistic sigmoid as $\sigma$ yields the proportional-odds (cumulative-logit) model, while using the standard normal CDF $\Phi$ instead recovers the probit link employed by GPOR. The per-class probabilities are the successive differences of these cumulative probabilities:

\begin{equation}
\hat{P}(y = k) =
\begin{cases}
    \sigma\bigl(\theta_1 - f(\bm{x})\bigr) & \text{if } k = 1 \\
    \sigma\bigl(\theta_k - f(\bm{x})\bigr) - \sigma\bigl(\theta_{k-1} - f(\bm{x})\bigr) & \text{if } 1 < k < K \\
    1 - \sigma\bigl(\theta_{K-1} - f(\bm{x})\bigr) & \text{if } k = K
\end{cases}
\label{equation:clm_probs}
\end{equation}

The ordering constraint $\theta_1 \leq \dots \leq \theta_{K-1}$ ensures that these differences are non-negative, so that Equation \ref{equation:clm_probs} defines a valid, rank-consistent probability distribution. For $K = 2$ it reduces to standard binary logistic regression.

The network parameters $\Theta$ and the cutpoints $\bm{\theta} = (\theta_1, \dots, \theta_{K-1})$ are learned jointly by minimizing the cumulative-link negative log-likelihood surrogate \cite{or_consistency} over the training set:

\begin{equation}
L(\Theta, \bm{\theta}) = -\sum_{i=1}^{N} \log \hat{P}(y = y_i \mid \bm{x}_i)
\label{equation:clm_loss}
\end{equation}

where $\hat{P}(y = y_i \mid \bm{x}_i)$ is given by Equation \ref{equation:clm_probs}. Because the cutpoints are learned directly from the data, the model adapts to non-uniform label distributions and spacings. At inference time, we predict the most probable class, $\hat{y} = \arg\max_{k} \hat{P}(y = k)$.

\subsubsection{NNRank}

NNRank \cite{cheng2008neural} recasts ordinal regression as multi-output binary classification. Each level is encoded as a binary target vector of length $K$: for sample $i$, the $k$-th entry is 1 whenever $k \leq y_i$ (Equation \ref{equation:nnrank_class}).

\begin{equation}
    t_{i,k} = \mathbf{1}\{k \leq y_i\}, \quad k = 1, \dots, K
\label{equation:nnrank_class}
\end{equation}

The output layer has $K$ neurons with sigmoid activations, so $f_k(\bm{x}_i) \approx P(y_i \geq k)$, and the loss is averaged over all $K$ outputs. This, together with the encoding, provides the ordinal inductive bias. Following the original work \cite{cheng2008neural}, we use the squared error as the base loss.
 
At inference time, each probability $f_k(\bm{x}_i)$ is thresholded at $0.5$, and the predicted level is the number of leading 1s (the count of positive outputs up to the first negative one). This does not guarantee consistency; for example, a pattern such as $[1, 1, 0, 1, 1, 1, 0, \dots]$ is possible. However, it arises rarely for well-performing models and did not appear in our initial experiments.

\subsubsection{OR-CNN}

The Ordinal Regression CNN (OR-CNN) \cite{niu2016ordinal} was originally proposed for age estimation. However, it uses a dedicated loss that can be used for any imbalanced ordinal regression problem. It transforms the $K$-class problem into $K-1$ binary tasks with the same encoding as NNRank, where $f_k(\bm{x}_i) \approx P(y_i > k)$. Unlike NNRank, the prediction is always the sum of the binary decisions of all tasks (Equation \ref{equation:orcnn_pred}).

\begin{equation}
\hat{y}_i = 1 + \sum_{k=1}^{K-1} \mathbf{1}\{ f_k(\bm{x}_i) > 0.5 \}
\label{equation:orcnn_pred}
\end{equation}

No consistency constraint is enforced between tasks, which simplifies training. To improve results under class imbalance, OR-CNN uses a weighted binary cross-entropy (rather than the squared error of NNRank), where the $k$-th task carries a weight $\lambda_k$ that is inversely related to $N_k$, the number of training samples for that task (Equation \ref{equation:orcnn_weight}).

\begin{equation}
    \lambda_k = \frac{\sqrt{N_k}}{\sum_{k'=1}^{K} \sqrt{N_{k'}}}
\label{equation:orcnn_weight}
\end{equation}

This gives higher weight to less populated tasks, and the authors show \cite{niu2016ordinal} that it improves overall results. The square root smooths the weights, avoiding excessively large weights for tasks with very few samples, which could cause the model to overfit them.

\subsubsection{CORAL}

The COnsistent RAnk Logits (CORAL) model \cite{coral2020} enforces prediction consistency, in contrast to the approaches above. Like OR-CNN, it decomposes the problem into $K-1$ binary tasks and predicts the level by summing the binary decisions $f_k(\bm{x}_i)$. CORAL enforces consistency by sharing the weight vector $\bm{w}$ of the output layer across all tasks, so that only the biases $b_k$ are task-specific.

Formally, consistency is expressed as rank monotonicity, $f_1(\bm{x}_i) \geq f_2(\bm{x}_i) \geq \dots \geq f_{K-1}(\bm{x}_i)$. The authors prove that minimizing the average binary cross-entropy over the $K-1$ tasks yields non-increasing biases $b_1 \geq b_2 \geq \dots \geq b_{K-1}$. Since $\bm{w}$ is shared, the outputs are then non-increasing as well, giving rank-monotonic and consistent predictions.

\subsubsection{CORN}

Conditional Ordinal Regression for Neural networks (CORN) \cite{corn} guarantees rank consistency without the weight-sharing constraint of CORAL, which can limit model expressiveness. Instead, CORN uses a training scheme based on conditional training subsets and the chain rule of probability. As in CORAL, the problem is decomposed into $K-1$ binary tasks, but CORN estimates a series of \emph{conditional} probabilities over the nested events $\{y_i > k\} \subseteq \{y_i > k-1\}$. The output of the $k$-th task is defined in Equation \ref{equation:corn_output}, and the unconditional probabilities are recovered by the chain rule in Equation \ref{equation:corn_prob}.

\begin{equation}
     f_k(\bm{x}_i) =
     \begin{cases}
         P(y_i > k \mid y_i > k-1) & \text{if } k > 1 \\
         P(y_i > 1) & \text{if } k = 1
     \end{cases}
\label{equation:corn_output}
\end{equation}
 
\begin{equation}
     P(y_i > k) = \prod_{j = 1}^{k} f_j(\bm{x}_i)
\label{equation:corn_prob}
\end{equation}
 
Because $0 \leq f_j(\bm{x}_i) \leq 1$ for all $j$, the unconditional probabilities are non-increasing, which guarantees rank consistency (Equation \ref{equation:corn_ranks}).

\begin{equation}
    P(y_i > 1) \geq P(y_i > 2) \geq \dots \geq P(y_i > K-1)
\label{equation:corn_ranks}
\end{equation}
 
Training uses conditional subsets, one per binary task. The subset $S_k$ is used to train $P(y_i > k \mid y_i > k-1)$ and contains only the samples that passed the previous threshold:
 
The loss minimized during backpropagation is
 
 \begin{equation}
     L(\Theta) = - \frac{\sum_{k = 1}^{K-1} \sum_{i \,:\, (\bm{x}_i, y_i) \in S_k} \Bigl[ \log f_k(\bm{x}_i)\, \mathbf{1}\{y_i > k\} + \log\bigl(1 - f_k(\bm{x}_i)\bigr)\, \mathbf{1}\{y_i \leq k\} \Bigr]}{\sum_{k=1}^{K-1} |S_k|}
 \end{equation}
 
For numerical stability, the authors also give an equivalent formulation in terms of logits $z_{i,k}$, where $\log f_k(\bm{x}_i) = \log \sigma(z_{i,k})$:
 
\begin{equation}
     L(\Theta) = - \frac{\sum_{k = 1}^{K-1} \sum_{i \,:\, (\bm{x}_i, y_i) \in S_k} \Bigl[ \log \sigma(z_{i,k})\, \mathbf{1}\{y_i > k\} + \bigl(\log \sigma(z_{i,k}) - z_{i,k}\bigr)\, \mathbf{1}\{y_i \leq k\} \Bigr]}{\sum_{k=1}^{K-1} |S_k|}
 \end{equation}
 
Prediction is as in CORAL: the predicted level is the count of unconditional probabilities of at least $0.5$. In CORN, however, rank consistency is guaranteed by construction of the training scheme.

\subsubsection{CONDOR}

CONDitionals for Ordinal Regression (CONDOR) \cite{condor} combines several of the advances above, providing built-in rank consistency without weight sharing. Like CORN, it transforms the task into $K-1$ binary problems, and like NNRank it uses the binary encoding $t_{i,k}$ of Equation \ref{equation:nnrank_class}. Predictions are obtained as in OR-CNN, by thresholding the marginal probabilities at $0.5$ and summing them (Equation \ref{equation:orcnn_pred}). As in CORN, CONDOR estimates the conditional probabilities $q_k(\bm{x}; \Theta)$ defined in Equation \ref{equation:condor_cond}.

\begin{equation}
    q_k(\bm{x}; \Theta) = P\bigl(y > k \mid y > k-1, \bm{x}; \Theta\bigr)
\label{equation:condor_cond}
\end{equation}

The boundary condition ``$y > 0$'' is set to unit probability. The marginal probabilities follow from the product rule, as in CORN (Equation \ref{equation:corn_prob}).

\begin{equation}
    p_k(\bm{x}; \Theta) = \prod_{j=1}^{k} q_j(\bm{x}; \Theta)
\label{equation:condor_marginal}
\end{equation}

Each output node represents one $q_k(\bm{x}; \Theta)$ and uses a sigmoid activation, so $q_k(\bm{x}; \Theta) \in (0,1)$ and the marginals of Equation \ref{equation:condor_marginal} are guaranteed to be non-increasing. Rank consistency thus holds by construction, for any upstream architecture. CONDOR is trained with a maximum likelihood loss. Letting $t_{i,k}$ be the binary target for task $k$ (Equation \ref{equation:nnrank_class}) and $\mathrm{BCE}(t, q) = -\bigl[t \log q + (1-t)\log(1-q)\bigr]$ the binary cross-entropy, the loss is defined in Equation \ref{equation:condor_loss}.

\begin{equation}
L(\Theta) = \sum_{i=1}^{N} \sum_{k=1}^{K-1} t_{i,k-1} \cdot \mathrm{BCE}\bigl(t_{i,k},\, q_k(\bm{x}_i; \Theta)\bigr)
\label{equation:condor_loss}
\end{equation}

The factor $t_{i,k-1}$ ensures that only the conditional probabilities on the ``active'' path (those with $t_{i,k-1} = 1$) contribute to the loss, mirroring the conditional subsets of CORN. The authors prove that minimizing this loss yields the maximum-likelihood estimate of $\Theta$. The key advantage of CONDOR is its universality: it guarantees rank-consistent outputs regardless of the upstream architecture, provided the final layer follows the CONDOR formulation. This makes it broadly applicable and theoretically robust compared with methods such as CORAL that require weight sharing.

\subsection{Train-test protocols}

To measure the generalization capabilities of the models, we need realistic evaluation protocols. A commonly used random split does not account for the structure or temporal dependencies in TTRPG data, as it does not capture the fact that monster designers often base new creations on existing ones. As a result, random splitting can introduce data leakage, where information from similar or related instances appears in both the training and test sets. This leads to an overly optimistic estimate of generalization performance, which undermines the reliability of the evaluation.

Instead, this setting more closely resembles time series data, where the newest samples should be used for evaluation. We therefore propose a \textit{chronological split}, which assigns older data to the training set and the newest monsters to the test set, based on the publication date of each book. The oldest monsters were published on 01.08.2019 and the newest ones included were published on 26.04.2026. The test set constitutes approximately $20\%$ of the data, and the oldest monsters in it were published on 01.08.2025. This approach reflects the natural emergence of new data over time and more accurately models how game designers would use the resulting models. A single chronological train-test split, however, assumes that we are interested in long-term predictions over large batches of monsters, since the test set covers samples far into the future relative to the training set.

We further propose a more refined strategy based on the domain's characteristics. Pathfinder monsters are naturally partitioned into chronologically ordered subsets, namely the books that contain them. During the design process, game designers focus on the next book or the few books scheduled for publication in the near future. Once those are published, the data can be added to the training set and the ML models retrained. This is equivalent to the \textit{expanding window} evaluation strategy from the time series forecasting literature. We first train on the initial portion of the data, test on the next batch, then add that batch to the training data, retrain the model, test on the next batch, and repeat until all available samples have been used. The reported score is the average metric over all test sets. Hyperparameters are tuned separately each time using cross-validation on the training data. While this procedure has a considerable computational cost, it yields a robust evaluation, reflecting real-world practice.

\begin{figure}
    \centering
    \includegraphics[width=\linewidth]{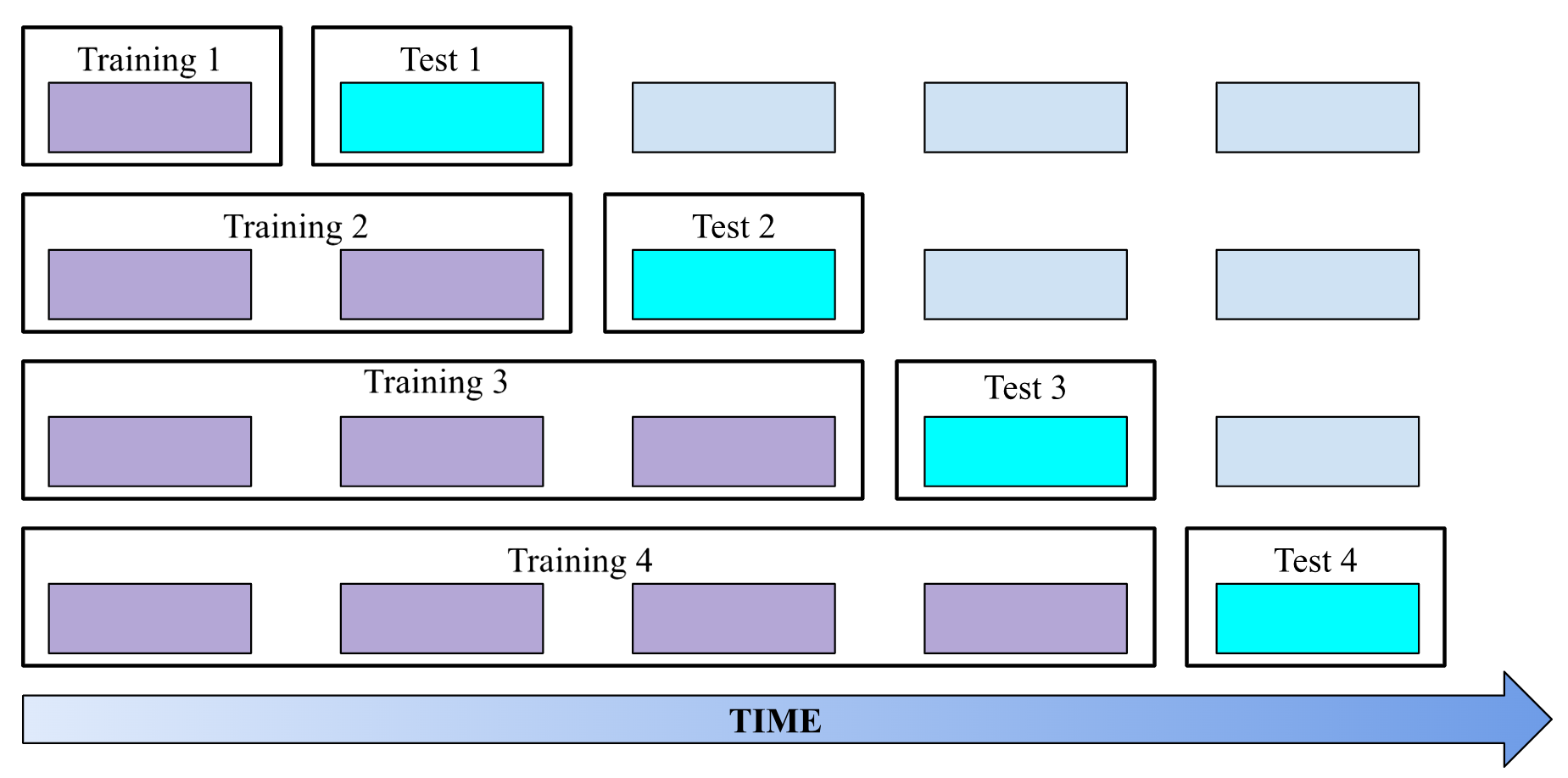}
    \caption{Expanding window evaluation.}
    \label{fig:expanding_window}
\end{figure}

In our case, the first subset includes all monsters published at the premiere of Pathfinder 2e: 7 sourcebooks with 232 monsters. We then test on subsequent books, ordered by publication date, ensuring that each test set contains at least 100 monsters, which lowers the variance of the estimate. To ensure temporal consistency, all monsters published on the same day were always assigned to the same subset. We visualize this in Figure \ref{fig:expanding_window}.

\subsection{Evaluation metrics}

The ordinal regression task combines aspects of both classification and standard regression. We therefore use a combination of metrics from both domains to comprehensively assess model performance, following best practices recommended in prior work \cite{baccianella2009evaluation}. We keep the notation of the previous subsections: $y_i$ is the true level of sample $i$, $\hat{y}_i$ the prediction, $K$ the number of classes, and $N_k$ the number of test samples of true class $k$.

To capture the magnitude of prediction errors, we use the standard RMSE and MAE. Because the dataset is imbalanced, with far more monsters at low levels than at high levels, these plain averages may not reflect prediction quality well. We therefore also use their macro-averaged versions $\mathrm{MAE}^M$ and $\mathrm{RMSE}^M$ \cite{baccianella2009evaluation}, which have been proposed specifically for imbalanced ordinal regression. They compute the metric separately for each class and then average over classes (Equations \ref{equation:rmse_m} and \ref{equation:mae_m}).

\begin{equation}
\mathrm{RMSE}^M = \frac{1}{K} \sum_{k=1}^{K} \sqrt{ \frac{1}{N_k} \sum_{i \,:\, y_i = k} \left(y_i - \hat{y}_i \right)^2 }
\label{equation:rmse_m}
\end{equation}
 
\begin{equation}
\mathrm{MAE}^M = \frac{1}{K} \sum_{k=1}^{K} \frac{1}{N_k} \sum_{i \,:\, y_i = k} \left| y_i - \hat{y}_i \right|
\label{equation:mae_m}
\end{equation}

Classification metrics assess the precision of the label predictions. Plain accuracy ($acc$) measures how often the prediction exactly matches the true label. It ignores error magnitude and is a harsh metric in ordinal regression. To contextualize performance, we also use accuracy-at-1 ($acc@1$), which counts a prediction one level above or below the ground truth as correct. It is always at least as large as plain accuracy.

To further assess the ordinal agreement between predicted and true labels, we use Somers' D rank correlation coefficient \cite{somers1962new}. It evaluates how often the predicted ordering of a pair of samples agrees with the ordering of their true labels: a pair is concordant if the two orderings match and discordant otherwise. Let $C$ and $D$ be the numbers of concordant and discordant pairs, and $T$ the number of pairs tied in the predicted scores. Somers' D is given in Equation \ref{equation:somers_D}; it lies in $[-1, 1]$, with $1$ indicating perfect agreement, $0$ no ordinal association, and $-1$ perfectly reversed ordering.

\begin{equation}
\text{Somers' } D (y, \hat{y}) = \frac{C - D}{C + D + T}
\label{equation:somers_D}
\end{equation}

\section{Experiments and results}

In this section, we present the experiments, results, and discussion. We start by analyzing the dataset and feature sets, and then validate the proposed evaluation scheme, namely chronological split and random split. Then, we present experiments with different rounding schemes, and finally a full comparison of regression with rounding, ordinal regression, and ordinal neural network models. Lastly, we present explainable AI analyses for the best performing model, as well as selected use cases.

\subsection{Experimental setup}

In every experiment, as the first step, features are normalized to the $[0, 1]$ range using min-max scaling, based on the training set. Unless stated otherwise, each experiment uses the chronological split.

In every experiment, hyperparameter optimization was performed using 5-fold cross-validation with MAE as the optimization criterion. Grid search was used for all models except LightGBM, which used the Tree-structured Parzen Estimator (TPE) \cite{tpe} due to its large number of hyperparameters. See Appendix \ref{appendix_hyperparameters} for hyperparameter grids.

Neural network models were trained with the AdamW optimizer for 40 epochs. Initial experiments indicated that this is enough for both training and validation losses to converge.

\subsection{Dataset analysis}

\begin{figure}[ht]
    \centering
    \includegraphics[width=\textwidth]{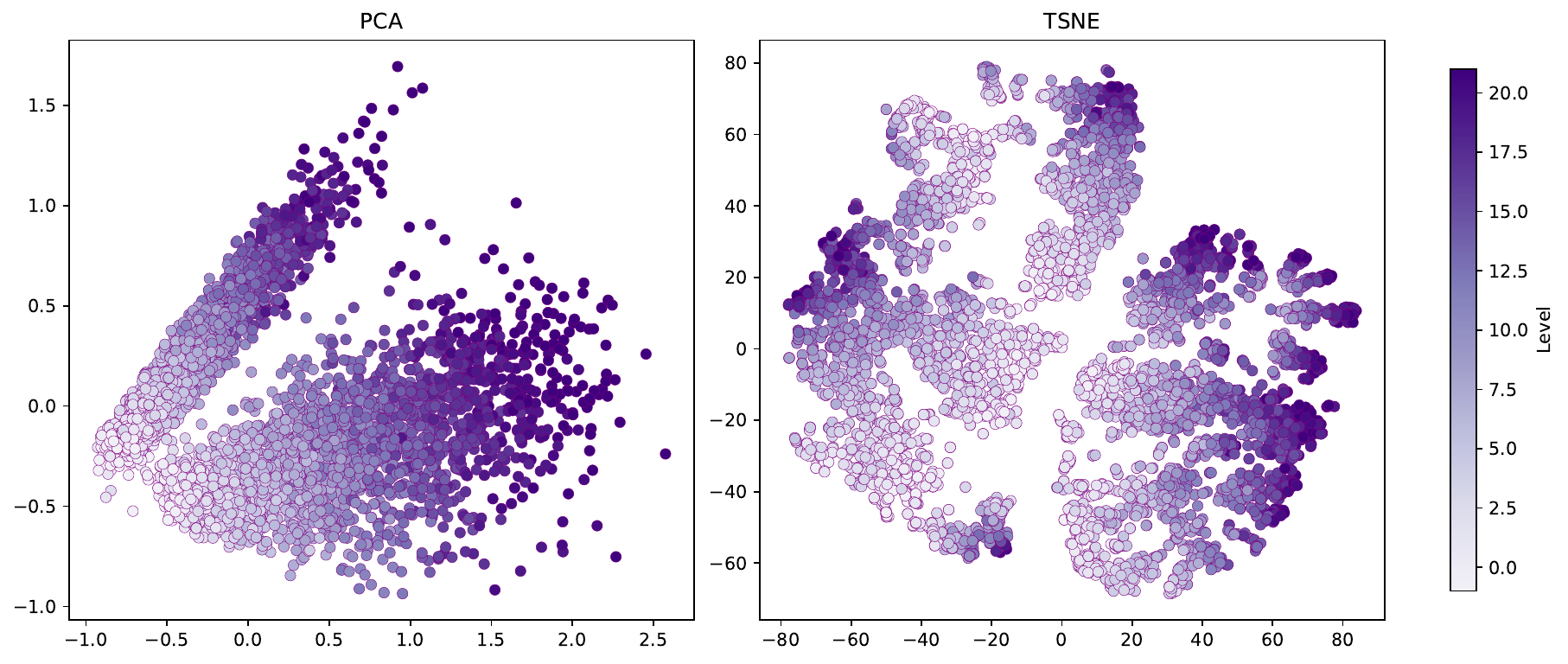}
    \caption{Dataset visualization after dimensionality reduction.}
    \label{fig:dim-red}
\end{figure}

For initial data inspection and visualization, we use principal component analysis (PCA) and t-SNE (default hyperparameters) using Expanded set of features, see Figure \ref{fig:dim-red}. Points have been colored based on their level, with lighter points indicating higher-level monsters. The dataset clearly shows a smooth transition of monster power, indicating well-engineered feature space. This also means that ML methods in the original space should have relatively good performance, as the feature-label relation should not be overly complex.

\begin{figure}[htp]
    \centering
    \includegraphics[width=\textwidth]{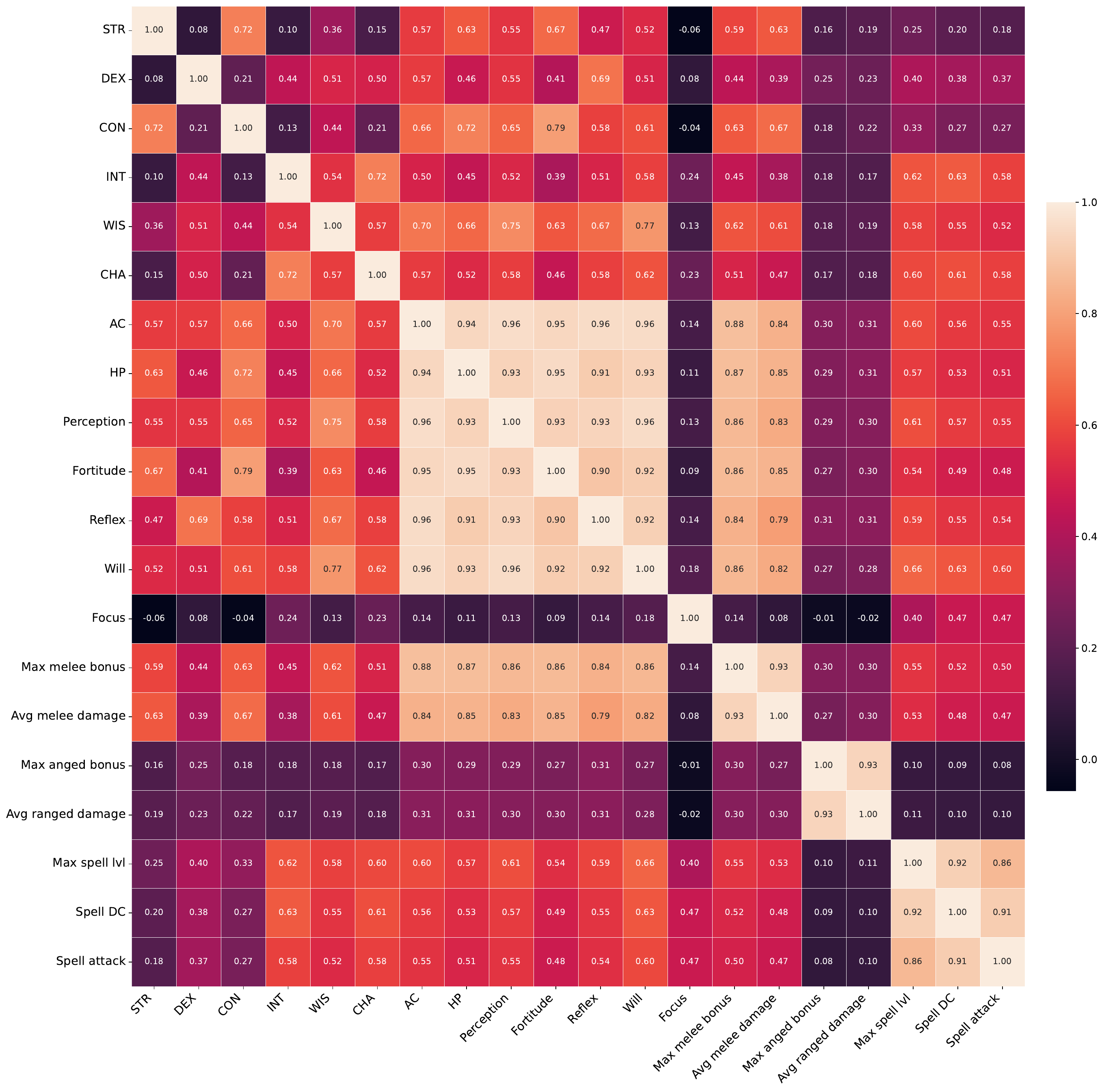}
    \caption{Pearson correlation between features.}
    \label{fig:correlation_matrix}
\end{figure}

Subsequently, we computed the Pearson correlation coefficients among all features using the Expa feature set, see Figure \ref{fig:correlation_matrix}. The resulting correlation matrix reveals several clusters of strongly intercorrelated variables. As anticipated, one such cluster consists of features characterizing essential spell-related attributes (e.g., spell attack bonus, maximum spell level, and spell save DC), which also exhibit strong correlations with more granular magic-related features, such as the number of spells per level. An additional set of highly correlated features is associated with defensive capabilities, including armor class (AC), hit points (HP), saving throws, and perception.

\begin{figure}[htp]
    \centering
    \includegraphics[width=0.8\textwidth]{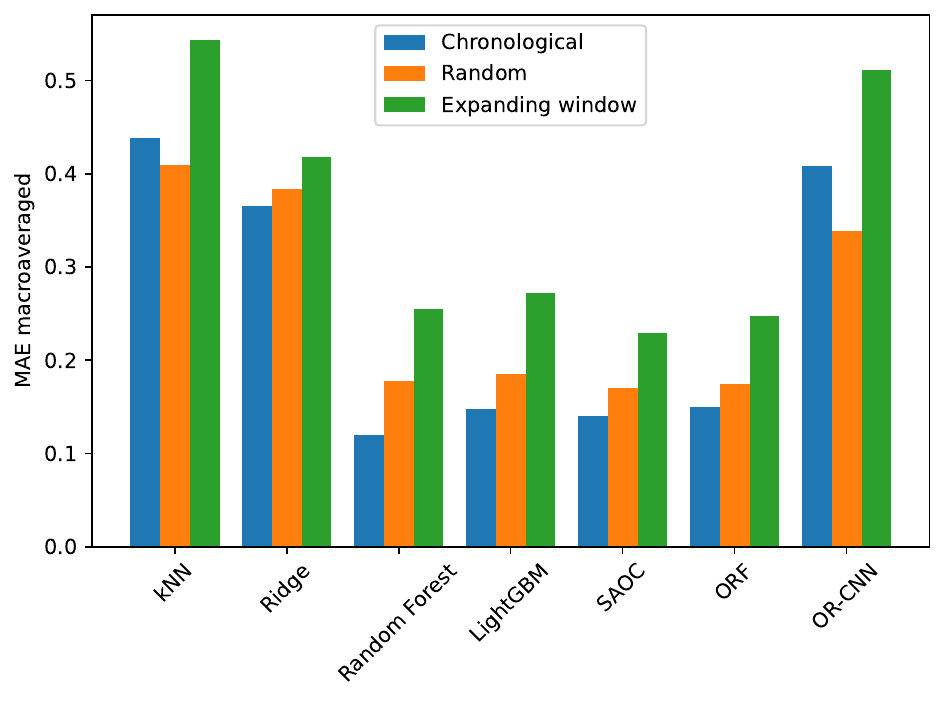}
    \caption{Performance of different evaluation strategies.}
    \label{fig:random_vs_chronological}
\end{figure}

In order to evaluate the usefulness of the proposed chronological split, we compare its results with random split on selected models. Interestingly, results in Figure \ref{fig:random_vs_chronological} show that random split often results in more pessimistic evaluation, i.e., higher macroaveraged MAE. This can be explained by the fact that designers often base new monsters on existing ones, which may may place similar samples in both the train and test sets. However, the expanding window strategy results in much more pessimistic evaluation in all cases, which may be caused by data distribution shifts, for example. As the chronological split and expanding window are more aligned with the design process, we focus on them in subsequent experiments.

Next, we evaluate the gain from incorporating more features, with detailed results provided in Appendix~\ref{appendix_feature_sets_comparison}. Among $16$ models evaluated, $12$ achieve results better or tied when using the Full feature set rather than Basic, with the other four being among the worst-performing models. Gains are often significant, on average lowering $MAE^M$ by about $0.10$.

However, using the Full feature set does not bring the same improvement, with most models performing very similarly, and $4$ achieving slightly better results. As such, we conclude that the Expanded feature set already appropriately represents the relationship between statistics and monster level. However, the models remain highly capable in case of more high-dimensional, richer representation space. As here we focus primarily on predictive performance, we use the Expanded feature set in subsequent experiments.

\subsection{Models performance comparison}

We present primary results of ML models for chronological split in Table \ref{tab:results_models_all}, and for expanding window in Table \ref{tab:results_window_models_all}. In all cases, we highlight the best model in bold, and second best in italics based on $MAE^M$. Based on those results and observations, we can draw several conclusions. They remain consistent to both chronological split and expanding window evaluation strategies.

\textbf{Conclusion 1: ML outperforms human-inspired baselines.} We included kNN models as baselines to model the human design process, which is based on comparison with existing monsters, as also recommended by the Pathfinder books. Their results are markedly inferior to those of all ML models, including even simple ridge regression. This difference is evident across all metrics, particularly in accuracy. This indicates the strong potential of ML to help human designers, and particularly amateur GMs, to create home-made monsters.

\textbf{Conclusion 2: Monster level prediction requires strong non-linear models.} The four best-performing models are tree-based ensembles: RF, LightGBM, ORF, and SAOC (with an underlying Random Forest). They strongly outperform linear models such as ridge regression, ORD, and Immediate-Threshold (IT). This indicates that monster level prediction is highly non-linear and requires sufficiently expressive models. Tree-based models also outperform the other non-linear models, including kernel SVM, GPOR, and neural networks. This can be explained by the limited data availability, which favors robust models that reduce error through ensembling.

\textbf{Conclusion 3: Neural networks for ordinal regression are not the best models for tabular data.} While all neural networks outperformed the baseline, none of those models matched the tree-based ensembles. They also showed higher standard deviations under expanding-window evaluation compared with other models, indicating relatively low stability and high variance. This can be attributed to the relatively small amount of data and aligns with the literature, which shows the superiority of tree-based ensembles over MLPs for tabular data in most cases. This holds despite the ordinal-regression-specific inductive biases incorporated into these neural models.

\textbf{Conclusion 4: Top models are strong and reliable.} The best-performing models - Random Forest, LightGBM, SAOC and ORF - achieved near-perfect Somers' D and acc@1. This means that they can almost always properly sort monsters by predicted power and their predictions nearly always lie within just 1 level of ground truth. Further, very low $MAE^M$, $RMSE^M$, and high accuracy indicate that predictions are exactly equal to the ground truth most of the time, and very close otherwise. The standard deviation of these models in case of expanding window is also relatively low across all metrics compared with the other models, meaning that they are stable and robust.

While there is still room for improvement, particularly in accuracy and $RMSE^M$, the created ML models are reliable tools for Pathfinder monster level prediction. The results achieved are good enough to warrant considering their deployment in practical applications supporting TTRPG game design.

\begin{table}[H]
  \centering
\begin{tabular}{|l||>{\centering\arraybackslash}m{2cm}|>{\centering\arraybackslash}m{2cm}|c|c|c|}
  \hline
    \textbf{Model} & \textbf{$\mathbf{MAE}^{\mathbf{M}}$} & \textbf{$\mathbf{RMSE}^{\mathbf{M}}$} & \textbf{Somers' D} & \textbf{Acc} & \textbf{Acc@1} \\
  \hline \hline
        kNN & 0.44 & 0.76 & 0.96 & 63\% & 96\% \\
        Ridge & 0.37 & 0.64 & 0.97 & 71\% & 98\% \\
        SVM & 0.21 & 0.49 & 0.98 & 83\% & 99\% \\
        \textbf{Random Forest} & \textbf{0.12} & \textbf{0.38} & 0.99 & \textbf{89\%} & 99\% \\
        \textit{LightGBM} & 0.14 & \textit{0.41} & 0.99 & 87\% & 99\% \\
      \hline
        SAOC & \textit{0.14} & 0.42 & 0.99 & \textit{88}\% & 99\% \\
        ORD & 0.32 & 0.61 & 0.97 & 74\% & 98\% \\
        IT & 0.31 & 0.60 & 0.97 & 75\% & 98\% \\
        ORF & 0.15 & 0.44 & 0.98 & 87\% & 99\% \\
        GPOR & 0.28 & 0.55 & 0.98 & 76\% & 99\% \\
      \hline
        CLM & 0.40 & 0.69 & 0.97 & 67\% & 97\% \\
        NNRank & 0.38 & 0.69 & 0.97 & 69\% & 97\% \\
        OR-CNN & 0.41 & 0.72 & 0.97 & 68\% & 97\% \\
        CORAL & 0.37 & 0.64 & 0.97 & 68\% & 98\% \\
        CORN & 0.25 & 0.53 & 0.98 & 78\% & 99\% \\
        CONDOR & 0.32 & 0.60 & 0.98 & 75\% & 98\% \\
      \hline
\end{tabular}
      \caption{Chronological split results. The best result marked in bold, second best in italics.}
      \label{tab:results_models_all}
\end{table}

\begin{table}[h]
    \centering
\begin{tabular}{|l||>{\centering\arraybackslash}m{2.3cm}|>{\centering\arraybackslash}m{2.3cm}|c|c|c|}
  \hline    
  \textbf{Model} & \textbf{$\mathbf{MAE}^{\mathbf{M}}$} & \textbf{$\mathbf{RMSE}^{\mathbf{M}}$} & \textbf{Somers' D} & \textbf{Acc} & \textbf{Acc@1} \\
  \hline
  \hline
        kNN & 0.54~$\pm$~0.17 & 0.91~$\pm$~0.22 & 0.94~$\pm$~0.01 & 58~$\pm$~8\% & 93~$\pm$~4\% \\
        Ridge & 0.42~$\pm$~0.13 & 0.78~$\pm$~0.21 & 0.96~$\pm$~0.01 & 68~$\pm$~6\% & 97~$\pm$~3\% \\
        SVM & 0.33~$\pm$~0.18 & 0.69~$\pm$~0.30 & 0.97~$\pm$~0.01 & 76~$\pm$~7\% & 97~$\pm$~3\% \\
        Random Forest & 0.25~$\pm$~0.12 & 0.57~$\pm$~0.19 & 0.97~$\pm$~0.01 & 79~$\pm$~8\% & 98~$\pm$~2\% \\
        \textit{LightGBM} & \textit{0.23~$\pm$~0.11} & 0.59~$\pm$~0.21 & 0.97~$\pm$~0.01 & 79~$\pm$~8\% & 98~$\pm$~2\% \\
      \hline
        \textbf{SAOC} & \textbf{0.23~$\pm$~0.09} & \textbf{0.53~$\pm$~0.11} & 0.97~$\pm$~0.01 & \textbf{81~$\pm$~7\%} & 98~$\pm$~2\% \\
        ORD & 0.37~$\pm$~0.12 & 0.73~$\pm$~0.17 & 0.96~$\pm$~0.01 & 71~$\pm$~6\% & 96~$\pm$~3\% \\
        IT & 0.36~$\pm$~0.11 & 0.70~$\pm$~0.17 & 0.96~$\pm$~0.01 & 72~$\pm$~6\% & 97~$\pm$~2\% \\
        ORF & 0.25~$\pm$~0.10 & \textit{0.56~$\pm$~0.14} & 0.97~$\pm$~0.01 & 80~$\pm$~7\% & 97~$\pm$~2\% \\
        GPOR & 0.34~$\pm$~0.14 & 0.67~$\pm$~0.20 & 0.97~$\pm$~0.01 & 72~$\pm$~6\% & 98~$\pm$~3\% \\
      \hline
        CLM & 0.66~$\pm$~0.70 & 1.07~$\pm$~0.85 & 0.93~$\pm$~0.06 & 58~$\pm$~12\% & 91~$\pm$~14\% \\
        NNRank & 0.55~$\pm$~0.25 & 0.91~$\pm$~0.33 & 0.96~$\pm$~0.02 & 59~$\pm$~12\% & 94~$\pm$~5\% \\
        OR-CNN & 0.51~$\pm$~0.22 & 0.87~$\pm$~0.30 & 0.96~$\pm$~0.02 & 63~$\pm$~10\% & 95~$\pm$~5\% \\
        CORAL & 0.53~$\pm$~0.29 & 0.89~$\pm$~0.37 & 0.96~$\pm$~0.02 & 60~$\pm$~14\% & 94~$\pm$~7\% \\
        CORN & 0.49~$\pm$~0.22 & 0.85~$\pm$~0.28 & 0.96~$\pm$~0.01 & 64~$\pm$~12\% & 95~$\pm$~5\% \\
        CONDOR & 0.59~$\pm$~0.46 & 0.96~$\pm$~0.55 & 0.96~$\pm$~0.02 & 64~$\pm$~10\% & 93~$\pm$~9\% \\
      \hline
\end{tabular}
\caption{Expanding window results. Mean and standard deviation over all test windows are reported. The best result (mean value) marked in bold, second best in italics.}
    \label{tab:results_window_models_all}
\end{table}

See Appendix \ref{appendix_metrics_reg_vs_macro} for comparison of regular and macroaveraged metrics. In Appendix \ref{appendix_expanding_window}, we also provide the detailed results for all testing windows under expanding window evaluation for the Random Forest.

\subsection{Best model analysis}

The results in the previous section show that many models perform very well on average. Here, we analyze the top-performing model on the chronological split, Random Forest, in more depth to inspect its behavior. We examine its error distribution, the relationship between error and out-of-distribution data, feature importances, and a few illustrative case studies. We use the chronological split throughout this section.

\begin{figure}[htp]
    \centering
    \includegraphics[width=0.7\textwidth]{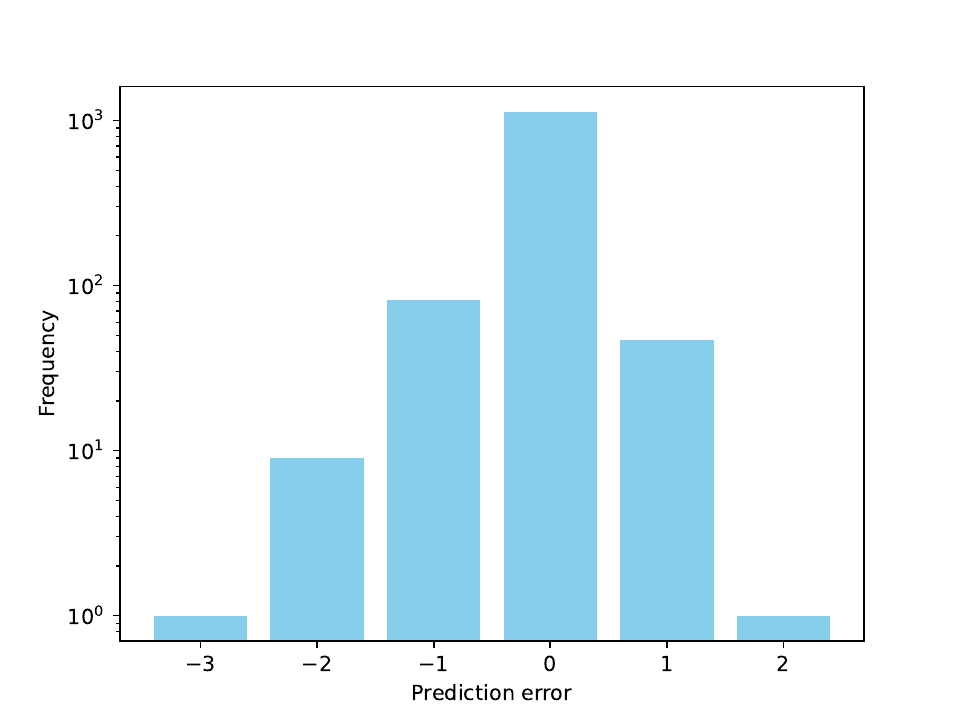}
    \caption{Errors histogram for Random Forest, logarithmic scale.}
    \label{fig:error-histogram}
\end{figure}

Figure \ref{fig:error-histogram} shows the distribution of prediction signed errors, i.e., $y - \hat{y}$, with logarithmic scale for readability. Predicting $1$ or $2$ levels too high (prediction error $-1$ or $-2$) is slightly more common, but this aligns with the imbalanced labels distribution in the dataset (see Figure \ref{fig:levels}). Most of the time, the model makes no mistake at all, or at most a single level, with just under $20$ cases having an error of $2$ or $3$ levels. As such, we can trust its predictions during the monster design process.

\begin{figure}[htp]
    \centering
    \includegraphics[width=0.7\textwidth]{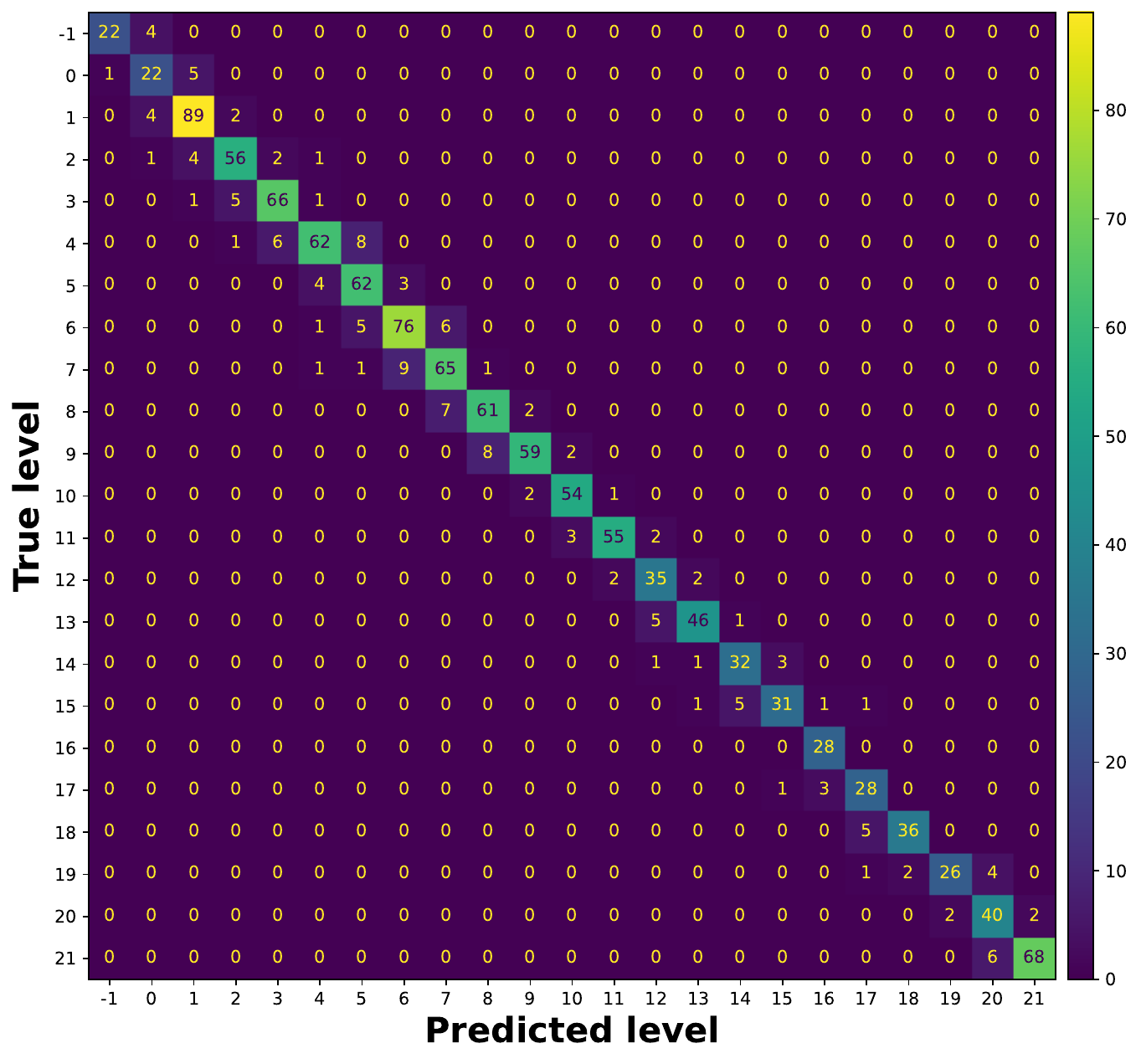}
    \caption{Confusion matrix for Random Forest.}
    \label{fig:results-cm-rf}
\end{figure}

The confusion matrix in Figure \ref{fig:results-cm-rf} confirms this: nearly all errors fall within a single level of the ground truth, and no systematic misclassification pattern emerges. Cases where the model makes a mistake may correspond to atypical monsters, for example those whose difficulty stems from unique special abilities rather than from their core statistics. Expanding the model to incorporate them is left as future work.

\begin{figure}[htp]
    \centering
    \includegraphics[width=0.7\textwidth]{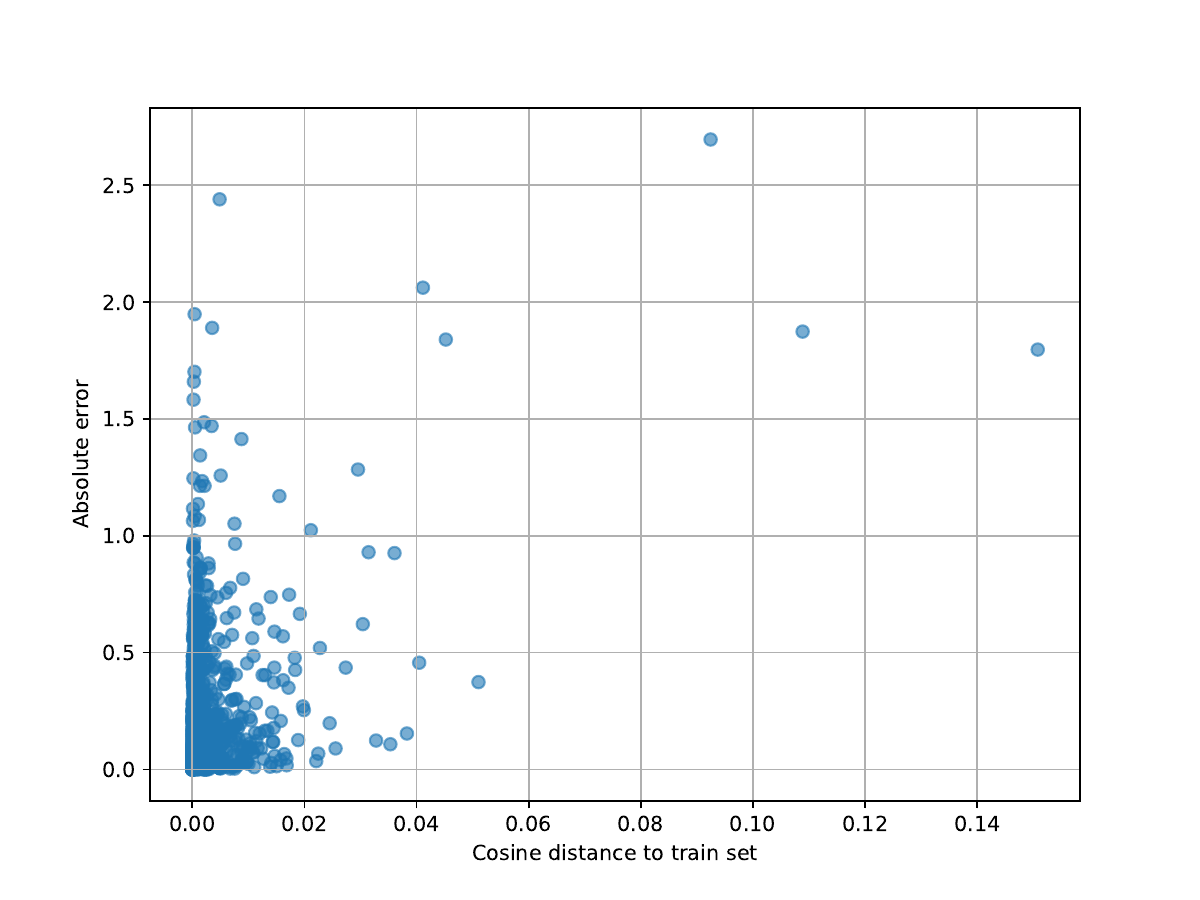}
    \caption{Distance to training data vs prediction error.}
    \label{fig:error-train-distance-distribution}
\end{figure}

Figure \ref{fig:error-train-distance-distribution} plots each test monster's absolute error against its cosine distance to the nearest training example. We observe no clear relationship between the two, and almost all distances are small. This suggests that new monsters rarely fall far from the training distribution and that the model is therefore not exposed to strongly out-of-distribution inputs.

\begin{figure}[htp]
    \centering
    \includegraphics[width=0.7\textwidth]{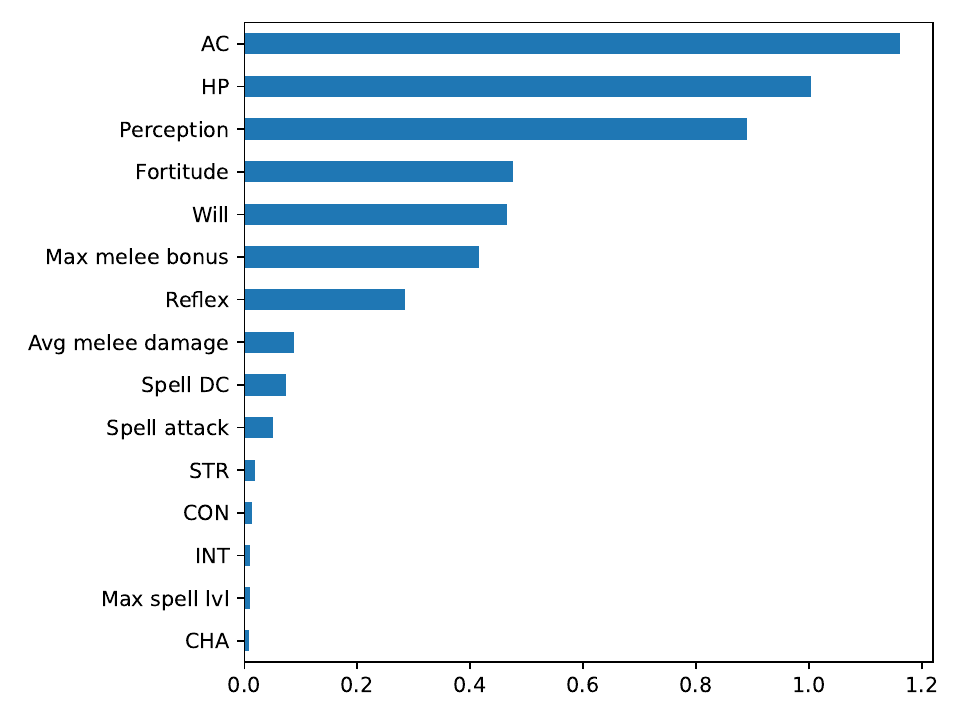}
    \caption{Mean absolute SHAP feature importances, top 15 features.}
    \label{fig:results-feature-importances}
\end{figure}

The analysis of feature importances, based on mean absolute SHAP values \cite{mean_absolute_shap,mean_absolute_shap_2}, is presented in Figure \ref{fig:results-feature-importances}. It indicates that the most influential predictors are primarily defensive statistics: Armor Class (AC), Hit Points (HP), Perception, and saving throws (Will, Fortitude, Reflex). Collectively, these variables capture a creature's ability to avoid or withstand damage, to detect threats early, and to influence the initiative order in combat.

Among offensive capabilities, the most important features are associated with melee combat, specifically the melee attack bonus and average melee damage. This aligns with the fact that melee is the most frequently used attack modality in the dataset. Spell-related statistics rank next in importance, reflecting their role as an alternative but still substantial source of offensive potential for magic-oriented creatures.

A notable finding is that features related to ranged attacks have comparatively low importance. They do not appear among top 15 most influential features. This also aligns with the fact that they are less common than melee and spell abilities among monsters in the dataset.

We conducted a case analysis of monsters with high prediction error in the test set. Most such cases correspond to entities designed for highly specific scenarios, frequently implemented as modified variants of well-known monsters. They possess the ``Unique'' tag and are highly customized for specific use cases, making them naturally challenging for ML. Another subset consists of creatures that are not independent monsters in the strict sense, but rather derivative or semi-autonomous creatures generated by another monster during combat. Although these derivatives are typically assigned the same difficulty level as their progenitor, they possess reduced statistics, which in practice leads to less challenging encounters. The only regular monster with a high prediction error is the Snowy Owl, whose statistics are more consistent with lower-level monsters. Its sole distinctive feature is flight, implemented as a special ability rather than as an inherent flying speed, which is not captured by the model. Extending featurization to include special abilities is left as future work.

In addition, we analyzed the model's performance for two common monster families, well represented both in training and test data: Trolls and Dragons. For the Troll family, the model showed no substantial difficulty in predicting the correct levels. The distance between the true class and the continuous output of the RF model tended to be slightly higher for cross-breed variants such as the Trollhund (a Troll-dog hybrid), yet the predictions remained close to the correct label, with an absolute error below $0.4$. Note that this family is represented in the test set by monsters with levels between $3$ and $15$.

The Dragon family is larger and predominantly composed of higher-level monsters. Individual dragons are typically represented by three age-based variants (Young, Adult, Ancient) and two role-based variants (standard and spellcaster). Most levels within this family were predicted correctly. We did not observe a consistent pattern in which the prediction error increases with the level (or age) of a given variant. No misclassifications occurred for Young dragons. In a few isolated instances an incorrect level was assigned. However, a misprediction for an Adult variant did not lead to an incorrect prediction for its corresponding Ancient variant. For the standard versus spellcaster forms, errors either occurred for both variants simultaneously or for neither.

\section{Conclusions}

This work investigated the application of machine learning to game design in tabletop role-playing games, focusing on monster level prediction in the Pathfinder Second Edition system. We created an openly available dataset, formalized the problem as ordinal regression, and validated our feature engineering through data analysis. We then compared a wide range of models - classical regression with rounding, tabular ordinal regression methods, and neural networks with ordinal-aware inductive biases - under multiple metrics and realistic validation strategies, including chronological and expanding-window evaluation.

Machine learning methods substantially outperformed human-inspired baselines, confirming their potential to assist both professional designers and amateur game masters. Tree-based ensembles, particularly Random Forest, LightGBM, ORF, and SAOC, achieved the best overall performance, capturing the highly non-linear feature-level relationships with accurate and stable predictions. Neural network approaches, including architectures designed for ordinal regression, did not outperform simpler models and showed higher variance. The best Random Forest model achieved near-perfect ranking quality and high accuracy, with errors typically within one level of the true value, and error analysis confirmed its robustness and alignment with human intuition. These results indicate that machine learning can serve as a reliable tool for monster design in Pathfinder 2e and, potentially, in similar rule-based games.

We plan to extend this research in several directions. First, designers may be interested not only in predicting a monster's level from its statistics, but also in actionable insights into how to change them to better match a target power. We plan to use counterfactual generation methods to propose minimal changes to a monster's statistics that gring it to the desired level, based on feature-rich Full set, providing data-driven expert-system support for game design. Second, since the target users are game designers without a technical background, we plan to conduct user studies to practically evaluate explainable AI methods for this problem, in order to understand the user experience aspects of post-hoc explanations of ordinal regression models.

\section*{Acknowledgements}

We thank the BIT Student Scientific Group for computational resources and administrative assistance.

\bibliographystyle{elsarticle-num} 
\bibliography{bibliography}

\appendix

\clearpage
\section{Feature sets comparison}
\label{appendix_feature_sets_comparison}

Here, we present the detailed $\mathrm{MAE}^M$ values for different feature sets on chronological split, see Table \ref{tab:appendix_full_vs_expanded}.

\begin{table}[H]
    \centering
\begin{tabular}{|l||c|c|c|c|c|}
  \hline
  \textbf{Model} & \textbf{Basic} & \textbf{Expanded} & \textbf{Full} & \textbf{\begin{tabular}[c]{@{}c@{}}Full better\\ than Basic\end{tabular}} & \textbf{\begin{tabular}[c]{@{}c@{}}Full better\\ than Expanded\end{tabular}} \\
  \hline
  \hline
        kNN & 0.64 & 0.44 & 0.54 & yes & no \\
        Ridge & 0.52 & 0.37 & 0.39 & yes & no \\
        SVM & 0.35 & 0.21 & 0.28 & yes & no \\
        Random Forest & 0.27 & 0.12 & 0.13 & yes & no \\
        LightGBM & 0.26 & 0.15 & 0.15 & yes & yes \\
        \hline
        SAOC & 0.26 & 0.14 & 0.16 & yes & no \\
        ORD & 0.47 & 0.32 & 0.34 & yes & no \\
        IT & 0.44 & 0.31 & 0.31 & yes & no \\
        ORF & 0.25 & 0.15 & 0.15 & yes & no \\
        GPOR & 0.36 & 0.28 & 0.34 & yes & no \\
        \hline
        CLM & 0.50 & 0.40 & 0.45 & yes & no \\
        NNRank & 0.53 & 0.38 & 0.60 & no & no \\
        OR-CNN & 0.39 & 0.41 & 0.41 & no & yes \\
        CORAL & 0.34 & 0.37 & 0.35 & no & yes \\
        CORN & 0.36 & 0.25 & 0.45 & no & no \\
        CONDOR & 0.42 & 0.32 & 0.31 & yes & yes \\
        \hline
    \hline
\end{tabular}
    \caption{Results of models on different feature sets.}
    \label{tab:appendix_full_vs_expanded}
\end{table}

\clearpage
\section{Regular vs macroaveraged metrics}
\label{appendix_metrics_reg_vs_macro}

In Tables \ref{tab:appendix_metrics_chronological} and \ref{tab:appendix_metrics_expanding_window}, we present the regular MAE and RMSE, as well as their macroaveraged variants. Macroaveraging always results in higher error values, often substantially. This follows the literature, confirming their usefulness for imbalanced ordinal regression.

\begin{table}[H]
    \centering
\begin{tabular}{|l||>{\centering\arraybackslash}m{2cm}|>{\centering\arraybackslash}m{2cm}|c|c|c|}
  \hline
  \textbf{Model} & \textbf{MAE} & \textbf{$\mathbf{MAE}^{\mathbf{M}}$} & \textbf{RMSE} & \textbf{$\mathbf{RMSE}^{\mathbf{M}}$} \\
  \hline
  \hline
        kNN & 0.42 & 0.44 & 0.74 & 0.76 \\
        Ridge & 0.31 & 0.37 & 0.59 & 0.64 \\
        SVM & 0.19 & 0.21 & 0.47 & 0.49 \\
        \textbf{Random Forest} & \textbf{0.12} & \textbf{0.12} & \textbf{0.37} & \textbf{0.38} \\
        LightGBM & 0.14 & 0.15 & \textit{0.41} & \textit{0.41} \\
      \hline
        \textit{SAOC} & \textit{0.13} & \textit{0.14} & 0.42 & 0.42 \\
        ORD & 0.29 & 0.32 & 0.59 & 0.61 \\
        IT & 0.27 & 0.31 & 0.57 & 0.60 \\
        ORF & 0.14 & 0.15 & 0.44 & 0.44 \\
        GPOR & 0.26 & 0.28 & 0.53 & 0.55 \\
      \hline
        CLM & 0.37 & 0.40 & 0.67 & 0.69 \\
        NNRank & 0.34 & 0.38 & 0.65 & 0.69 \\
        OR-CNN & 0.36 & 0.41 & 0.68 & 0.72 \\
        CORAL & 0.33 & 0.37 & 0.61 & 0.64 \\
        CORN & 0.24 & 0.25 & 0.52 & 0.53 \\
        CONDOR & 0.26 & 0.32 & 0.54 & 0.60 \\
    \hline
\end{tabular}
    \caption{Unweighted and macroaveraged metrics, chronological split.}
    \label{tab:appendix_metrics_chronological}
\end{table}

\begin{table}[h]
    \centering
\begin{tabular}{|l||>{\centering\arraybackslash}m{2.3cm}|>{\centering\arraybackslash}m{2.3cm}|c|c|c|}
  \hline
  \textbf{Model} & \textbf{MAE} & \textbf{$\mathbf{MAE}^{\mathbf{M}}$} & \textbf{RMSE} & \textbf{$\mathbf{RMSE}^{\mathbf{M}}$} \\
  \hline
  \hline
        kNN & 0.50~$\pm$~0.14 & 0.54~$\pm$~0.17 & 0.84~$\pm$~0.18 & 0.91~$\pm$~0.22 \\
        Ridge & 0.37~$\pm$~0.09 & 0.42~$\pm$~0.13 & 0.73~$\pm$~0.17 & 0.78~$\pm$~0.21 \\
        SVM & 0.29~$\pm$~0.15 & 0.33~$\pm$~0.18 & 0.63~$\pm$~0.26 & 0.69~$\pm$~0.30 \\
        \textit{Random Forest} & \textit{0.23~$\pm$~0.10} & \textit{0.25~$\pm$~0.12} & \textit{0.53~$\pm$~0.15} & \textit{0.57~$\pm$~0.19} \\
        LightGBM & 0.24~$\pm$~0.10 & 0.27~$\pm$~0.13 & 0.55~$\pm$~0.15 & 0.60~$\pm$~0.22 \\
      \hline
        \textbf{SAOC} & \textbf{0.22~$\pm$~0.09} & \textbf{0.23~$\pm$~0.09} & \textbf{0.51~$\pm$~0.11} & \textbf{0.53~$\pm$~0.11} \\
        ORD & 0.34~$\pm$~0.08 & 0.37~$\pm$~0.12 & 0.69~$\pm$~0.14 & 0.73~$\pm$~0.17 \\
        IT & 0.33~$\pm$~0.08 & 0.36~$\pm$~0.11 & 0.67~$\pm$~0.14 & 0.70~$\pm$~0.17 \\
        ORF & \textit{0.23~$\pm$~0.09} & \textit{0.25~$\pm$~0.10} & 0.54~$\pm$~0.13 & 0.56~$\pm$~0.14 \\
        GPOR & 0.31~$\pm$~0.09 & 0.34~$\pm$~0.14 & 0.64~$\pm$~0.16 & 0.67~$\pm$~0.20 \\
      \hline
        CLM & 0.60~$\pm$~0.57 & 0.66~$\pm$~0.70 & 0.98~$\pm$~0.70 & 1.07~$\pm$~0.85 \\
        NNRank & 0.50~$\pm$~0.20 & 0.55~$\pm$~0.25 & 0.84~$\pm$~0.29 & 0.91~$\pm$~0.33 \\
        OR-CNN & 0.44~$\pm$~0.18 & 0.51~$\pm$~0.22 & 0.80~$\pm$~0.27 & 0.87~$\pm$~0.30 \\
        CORAL & 0.48~$\pm$~0.24 & 0.53~$\pm$~0.29 & 0.82~$\pm$~0.30 & 0.89~$\pm$~0.37 \\
        CORN & 0.43~$\pm$~0.19 & 0.49~$\pm$~0.22 & 0.77~$\pm$~0.25 & 0.85~$\pm$~0.28 \\
        CONDOR & 0.48~$\pm$~0.32 & 0.59~$\pm$~0.46 & 0.85~$\pm$~0.44 & 0.96~$\pm$~0.55 \\
        \hline
    \hline
\end{tabular}
    \caption{Unweighted and macroaveraged metrics, expanding window.}
    \label{tab:appendix_metrics_expanding_window}
\end{table}

\clearpage
\section{Detailed expanding window results}
\label{appendix_expanding_window}

In Table \ref{tab:appendix_expanding_window}, we provide the detailed results over all testing windows under the expanding window evaluation for Random Forest. Following the protocol explained in the main body, we obtained 21 train-test runs (windows). The metrics remain generally consistent after a few runs, with highly noisy, but overall accurate results. This can be explained by data distribution shifts as novel monsters are introduced.

\begin{table}[h]
    \centering
\begin{tabular}{|l||>{\centering\arraybackslash}m{2cm}|>{\centering\arraybackslash}m{2cm}|c|c|c|}
  \hline    
  \textbf{Run} & \textbf{$\mathbf{MAE}^{\mathbf{M}}$} & \textbf{$\mathbf{RMSE}^{\mathbf{M}}$} & \textbf{Somers' D} & \textbf{Acc} & \textbf{Acc@1} \\
  \hline
  \hline
        1 & 0.61 & 1.08 & 0.96 & 58\% & 93\% \\
        2 & 0.44 & 0.76 & 0.96 & 71\% & 97\% \\
        3 & 0.20 & 0.44 & 0.99 & 84\% & 100\% \\
        4 & 0.19 & 0.46 & 0.97 & 81\% & 100\% \\
        5 & 0.26 & 0.63 & 0.97 & 78\% & 98\% \\
        6 & 0.47 & 0.94 & 0.96 & 71\% & 95\% \\
        7 & 0.32 & 0.65 & 0.98 & 65\% & 99\% \\
        8 & 0.22 & 0.49 & 0.98 & 83\% & 99\% \\
        9 & 0.20 & 0.49 & 0.99 & 86\% & 99\% \\
        10 & 0.12 & 0.38 & 0.98 & 85\% & 100\% \\
        11 & 0.26 & 0.63 & 0.97 & 76\% & 96\% \\
        12 & 0.24 & 0.57 & 0.98 & 80\% & 98\% \\
        13 & 0.22 & 0.49 & 0.98 & 85\% & 98\% \\
        14 & 0.27 & 0.75 & 0.94 & 73\% & 97\% \\
        15 & 0.26 & 0.51 & 0.98 & 85\% & 100\% \\
        16 & 0.24 & 0.57 & 0.98 & 77\% & 98\% \\
        17 & 0.17 & 0.42 & 0.98 & 87\% & 100\% \\
        18 & 0.20 & 0.48 & 0.98 & 84\% & 99\% \\
        19 & 0.25 & 0.59 & 0.97 & 80\% & 98\% \\
        20 & 0.12 & 0.35 & 0.97 & 85\% & 99\% \\
        21 & 0.07 & 0.27 & 0.99 & 93\% & 100\% \\
        \hline
      \hline
\end{tabular}
\caption{Detailed expanding window results for Random Forest.}
\label{tab:appendix_expanding_window}
\end{table}

\clearpage
\section{Hyperparameter grids}
\label{appendix_hyperparameters}

Here, we present the hyperparameter grids evaluated in our experiments. Regression and tabular ordinal regression models' grids are summarized in Table \ref{tab:hyperparams_classical}. $d$ denotes the number of input features. All neural networks used the same grid: learning rate $\{10^{-3}, 10^{-2}, 10^{-1}\}$, weight decay $\{10^{-3}, 10^{-2}, 10^{-1}, 1\}$.

Grid search with 5-fold cross-validation was used to search through all grids, except for LightGBM. In that case, due to large hyperparameter space, Tree Parzen Estimator (TPE) \cite{tpe} was used, as implemented in the Optuna library \cite{optuna}. We used 100 trials for tuning, with 10 random search trials for warmup, considering the Gaussian prior and using multivariate TPE. In the table, we indicate the distributions for sampling hyperparameters, with minimal and maximal values for each.

\begin{table}[h]
\centering
\small
\begin{tabular}{|l|l|l|}
\hline
\textbf{Model} & \textbf{Hyperparameter} & \textbf{Grid / value} \\
\hline
\hline
Ridge & $\alpha$ & 10\,000 values, linearly spaced in $[10^{-3}, 1]$ \\
\hline
SVM (RBF kernel) & $C$ & 100 values, linearly spaced in $[1, 10]$ \\
\hline
\multirow{4}{*}{kNN}
    & \texttt{n\_neighbors} & $\{1, 3\}$ \\
    & \texttt{weights} & $\{$uniform, distance$\}$ \\
    & \texttt{metric} & $\{$Minkowski, Manhattan, Euclidean$\}$ \\
    & \texttt{leaf\_size} & $\{50, 60, 70, 80, 90\}$ \\
\hline
\multirow{3}{*}{Random Forest}
    & \texttt{max\_features} & $\{\sqrt{d},\ 0.3d\}$ \\
    & \texttt{n\_estimators} & $\{100, 200, 500\}$ \\
    & \texttt{criterion} & $\{$squared error, absolute error, Friedman MSE$\}$ \\
\hline
\multirow{5}{*}{LightGBM}
    & \texttt{n\_estimators} & integer log-uniform ($100$, $1000$) \\
    & \texttt{num\_leaves} & integer uniform ($15$, $127$) \\
    & \texttt{min\_child\_samples} & integer uniform ($5$, $100$) \\
    & \texttt{learning\_rate} & float log-uniform ($0.01$, $0.1$) \\
    & \texttt{colsample\_bytree} & float uniform ($0.5$, $1.0$) \\
    & \texttt{reg\_lambda} & float log-uniform ($0.001$, $10.0$) \\
\hline
\multirow{4}{*}{ORF}
    & \texttt{max\_features} & $\{0.3d\}$ \\
    & \texttt{min\_samples\_leaf} & $\{2, 3, 4, 5, 6, 7\}$ \\
    & \texttt{n\_estimators} & $\{100, 200, 500\}$ \\
    & \texttt{honesty} & False \\
\hline
IT & $\alpha$ & 100 values, linearly spaced in $[0, 10^{-3}]$ \\
\hline
ORD & offset & 11 values, linearly spaced in $[0.25, 1.25]$ \\
\hline
\multirow{3}{*}{SAOC (Random Forest)}
    & \texttt{max\_features} & $\{\sqrt{d},\ 0.3d\}$ \\
    & \texttt{n\_estimators} & $\{100, 200, 500\}$ \\
    & \texttt{criterion} & $\{$Gini, entropy$\}$ \\
\hline
\multirow{2}{*}{GPOR}
    & kernel & ArcCosine \\
    & \texttt{max\_iter} & 100 \\
\hline
\end{tabular}
\caption{Hyperparameter grids for regression and tabular ordinal regression models.}
\label{tab:hyperparams_classical}
\end{table}

\end{document}